\documentclass[runningheads]{llncs}

\usepackage{eccv}

\usepackage{eccvabbrv}

\usepackage{graphicx}
\usepackage{booktabs}

\usepackage[accsupp]{axessibility}  

\usepackage[breaklinks,colorlinks,allcolors=black,citecolor=eccvblue]{hyperref}

\usepackage{orcidlink}

\usepackage{booktabs}
\usepackage{multirow}
\usepackage{colortbl}
\usepackage{xcolor}
\usepackage{adjustbox}
\usepackage[most]{tcolorbox}
\tcbuselibrary{listings} 

\usepackage{wrapfig}
\usepackage{multicol}
\usepackage{tikz}
\usepackage{contour}
\usepackage{graphicx}

\definecolor{c1}{HTML}{E6324C}
\definecolor{c2}{HTML}{6EC5C1}
\definecolor{c3}{HTML}{8A7CCE}
\definecolor{c4}{HTML}{7E2EA2}

\usepackage{subcaption}
\usepackage[table]{xcolor}
\usepackage{graphicx}
\usepackage{tabularx}
\usepackage{array}
\usepackage{ragged2e}
\usepackage{longtable}
\usepackage[table]{xcolor}
\usepackage{tabularx}
\usepackage{booktabs}
\usepackage{array}
\usepackage{ragged2e}
\usepackage{caption}

\usepackage{appendix}
\usepackage{float}

\usepackage{titletoc}

\definecolor{greenheader}{HTML}{D9EAD3}
\definecolor{blueheader}{HTML}{CFE2F3}
\definecolor{yellowheader}{HTML}{FFF2CC}
\definecolor{purpleheader}{HTML}{D9D2E9}
\definecolor{orangeheader}{HTML}{FCE5CD}

\definecolor{bestgreen}{HTML}{C8E6C9}

\newtcbox{\linkbadge}{on line, boxsep=2pt, left=5pt, right=5pt, top=2pt, bottom=2pt,
  colframe=black!55, colback=black!3, boxrule=0.5pt, arc=4pt}

\makeatletter
\newcommand{\printfnsymbol}[1]{%
  \textsuperscript{\@fnsymbol{#1}}%
}
\makeatother

\usepackage{fontawesome5}

\definecolor{navylink}{HTML}{1F3A5F}

\begin{document}

\title{MultihopSpatial: Multi-hop Compositional Spatial Reasoning Benchmark for Vision-Language Model} %

\titlerunning{MultihopSpatial}


\author{Youngwan Lee\thanks{Equal contribution}\inst{1,2}\orcidlink{0000-0001-8644-155X} \and
Soojin Jang\printfnsymbol{1}\inst{1}\orcidlink{0000-0002-2719-7646} \and
Yoorhim Cho\inst{1}\orcidlink{0009-0004-7179-9291} \and
Seunghwan Lee\inst{3}\orcidlink{0009-0006-5762-4107} \and \\
Yong-Ju Lee\inst{1}\orcidlink{0000-0001-6538-4701} \and
Sung Ju Hwang\inst{2,4}\orcidlink{0000-0002-9675-2324}} 

\authorrunning{Y. Lee et al.}

\institute{Electronics and Telecommunications Research Institute, South Korea \and
Korea Advanced Institute of Science and Technology, South Korea \and
Sungkyunkwan University, South Korea \and
DeepAuto, South Korea
}

\maketitle

{\hypersetup{urlcolor=navylink}
\begin{center}
\href{https://youngwanlee.github.io/multihopspatial/}{\color{navylink}\faGlobe~Website}\hspace{1.2em}
\href{https://huggingface.co/datasets/etri-vilab/MultihopSpatial}{%
  \raisebox{-0.2em}{\includegraphics[height=1em]{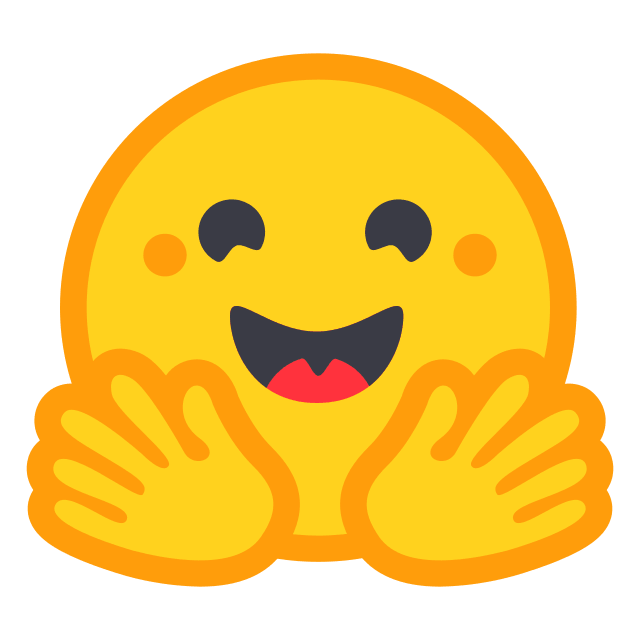}}\color{navylink}~Dataset}\hspace{1.2em}
\href{https://huggingface.co/etri-vilab/MultiHopSpatial-Qwen3-VL-4B-Instruct}{%
  \raisebox{-0.2em}{\includegraphics[height=1em]{fig/huggingface-color.png}}\color{navylink}~Model}\hspace{1.2em}
\href{https://github.com/youngwanLEE/multihopspatial}{\color{navylink}\faGithub~Code}
\end{center}
}

\begin{abstract} 
Spatial reasoning is foundational for Vision-Language Models (VLMs), particularly when deployed as Vision-Language-Action (VLA) agents in physical environments. However, existing benchmarks predominantly focus on elementary, single-hop relations, neglecting the multi-hop compositional reasoning and precise visual grounding essential for real-world scenarios. To address this, we introduce \textbf{MultihopSpatial}, offering three key contributions: (1) A comprehensive benchmark designed for multi-hop and compositional spatial reasoning, featuring 1- to 3-hop complex queries across diverse spatial perspectives. (2) \textbf{Acc@50IoU}, a complementary metric that simultaneously evaluates reasoning and visual grounding by requiring both answer selection and precise bounding box prediction—capabilities vital for robust VLA deployment. (3) \textbf{MultihopSpatial-Train}, a dedicated large-scale training corpus to foster spatial intelligence. Extensive evaluation of 37 state-of-the-art VLMs yields eight key insights, revealing that compositional spatial reasoning remains a formidable challenge. Finally, we demonstrate that reinforcement learning post-training on our corpus enhances both intrinsic VLM spatial reasoning and downstream embodied manipulation performance.
\end{abstract}
\section{Introduction}\label{sec:intro}
\begin{figure}[t]
    \centering
        \includegraphics[width=\linewidth]{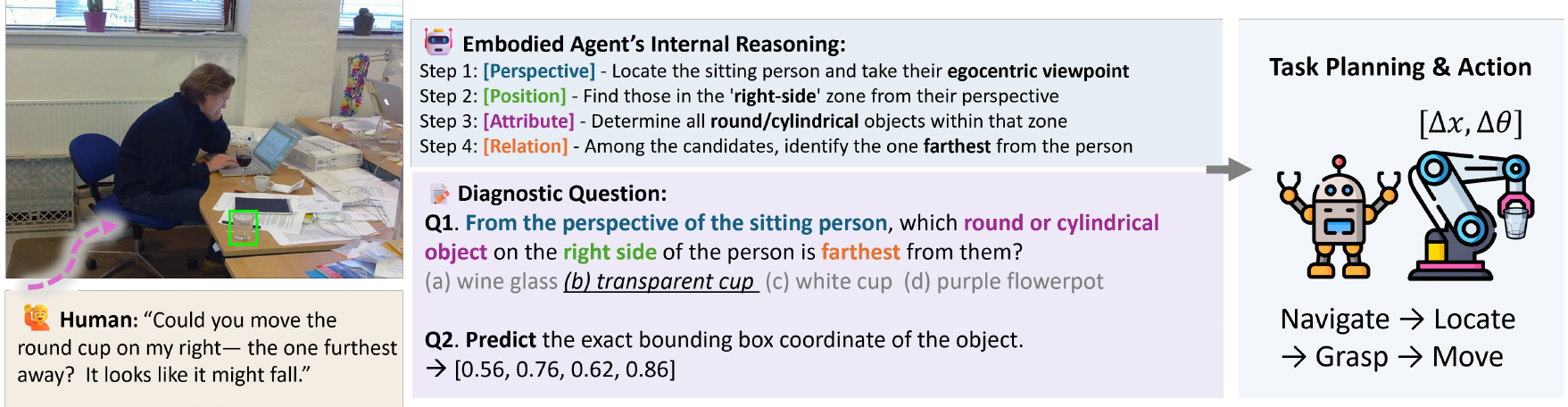}
    \vspace{-0.5cm}
    \caption{\textbf{A multi-hop spatial reasoning example for an embodied agent in the real-world scenario.}}
    \vspace{-0.5cm}
    \label{fig:teaser_1}
\end{figure}

The recent surge of interest in physical AI has accelerated the development of embodied agents, particularly Vision-Language-Action~(VLA) models~\cite{brohan2023rt2, kim2024openvla,yang2025vlaser,li2025physiagent}. These agents fundamentally rely on Vision-Language Models~(VLMs) for spatial reasoning to interact with the real world. However, existing VLMs often lack precise visual grounding, leading to significant difficulties in accurate perception and action execution within VLA frameworks~\cite{zhang2026vlmvla, zhang2026falcon, huang2025roboground, chen2025goal}. In complex environments, agents must execute multi-step, compositional spatial reasoning and accurately localize targets to succeed. As illustrated in~\cref{fig:teaser_1}, an instruction like ``\textit{Could you move the round cup on my right---the one furthest away?}'' requires an agent to adopt an ego-centric perspective, isolate the target position, filter by attributes, and compare relations. This internal reasoning perfectly mirrors a diagnostic multi-hop multiple-choice question (MCQ) coupled with exact bounding box prediction. Ultimately, an agent can only successfully navigate and manipulate an object if it accurately answers the query and visually grounds the target.

Despite rapid advancements, existing spatial reasoning benchmarks largely focus on single-hop queries, such as verifying elementary relations like "Is X to the right of Y?"~\cite{spatialmqa, EmbSpatial, 3dsrbench, omnispatial25, spatialab}. As depicted in the left panel of~\cref{fig:teaser_2}, these datasets typically evaluate spatial understanding through standard MCQs without requiring spatial localization. This design not only fails to capture the compositional reasoning required in dynamic physical environments but also leaves a significant spatial blind spot in VLM evaluation~\cite{alam2026blindspot}, as models can often select the correct answer without genuinely locating the target. Furthermore, while some recent efforts provide training sets alongside their benchmarks~\cite{spatialscore,omnispatial25}, they typically do not extend their validation to end-to-end VLA execution, leaving a profound gap between VLM spatial reasoning metrics and actual embodied action performance.

To bridge this gap, we introduce \textbf{MultihopSpatial}, a comprehensive benchmark evaluating multi-hop, compositional spatial reasoning paired with visual grounding. It comprises 4,500 QA pairs spanning 1- to 3-hop complexities across attribute, position, and relation dimensions, encompassing both ego-centric and exo-centric viewpoints to mirror real-world interactions. To ensure robust evaluation for VLA deployment, we introduce a complementary grounded metric, \textbf{Acc@50IoU}, which considers an answer correct only if the predicted bounding box overlaps the ground-truth by at least 50\% IoU. The critical necessity of this metric is starkly illustrated in~\cref{fig:teaser_2} (right): while frontier VLMs achieve superficially high standard MCQ accuracy (solid bars), their performance plummets under Acc@50IoU (striped bars). This severe discrepancy confirms that standard MCQs mask a profound lack of spatial grounding, validating Acc@50IoU as an essential tool to expose genuine capabilities. Finally, we provide a 6,791-sample training set, validating the dataset's utility not merely as an evaluation tool, but as a potent corpus for enhancing downstream VLA performance.

We extensively evaluate 37 VLMs, encompassing state-of-the-art commercial, open-weight, and specialized spatial reasoning models. Our findings reveal that multi-hop spatial reasoning remains a formidable challenge; for instance, a highly capable reasoning model (\eg, GPT-5.2-Thinking) drops from 45.8\% in standard MCQ to a mere 9.4\% under our grounded metric on MultihopSpatial 3-Hop in~\cref{fig:teaser_2}~(right). Beyond evaluation, we leverage the MultihopSpatial-Train set to post-train a base VLM using reinforcement learning~(\eg, GRPO~\cite{deepseekmath}). 
Our experiments demonstrate that this RL post-training not only enhances the model's intrinsic multi-hop spatial reasoning capabilities across five benchmarks but also translates to improved performance in two downstream embodied VLA tasks.

Our main contributions are summarized as follows:
\begin{itemize}
    \item \textbf{MultihopSpatial:} A novel benchmark jointly evaluating multi-hop, compositional spatial reasoning and visual grounding in VLMs.
    \item \textbf{Acc@50IoU:} A complementary grounded metric requiring both correct answer selection and precise bounding box prediction, eliminating the blind spot of traditional MCQs.
    \item \textbf{MultihopSpatial-Train:} A dedicated training corpus where RL post-training enhances both intrinsic VLM spatial intelligence and downstream VLA manipulation performance.
\end{itemize}
\section{Related Work}

\definecolor{persp}{HTML}{156082}  
\definecolor{attri}{HTML}{A02B93}  
\definecolor{posit}{HTML}{4EA72E}  
\definecolor{relat}{HTML}{E97132}  

\begin{figure}[t]
    \centering
        \includegraphics[width=\linewidth]{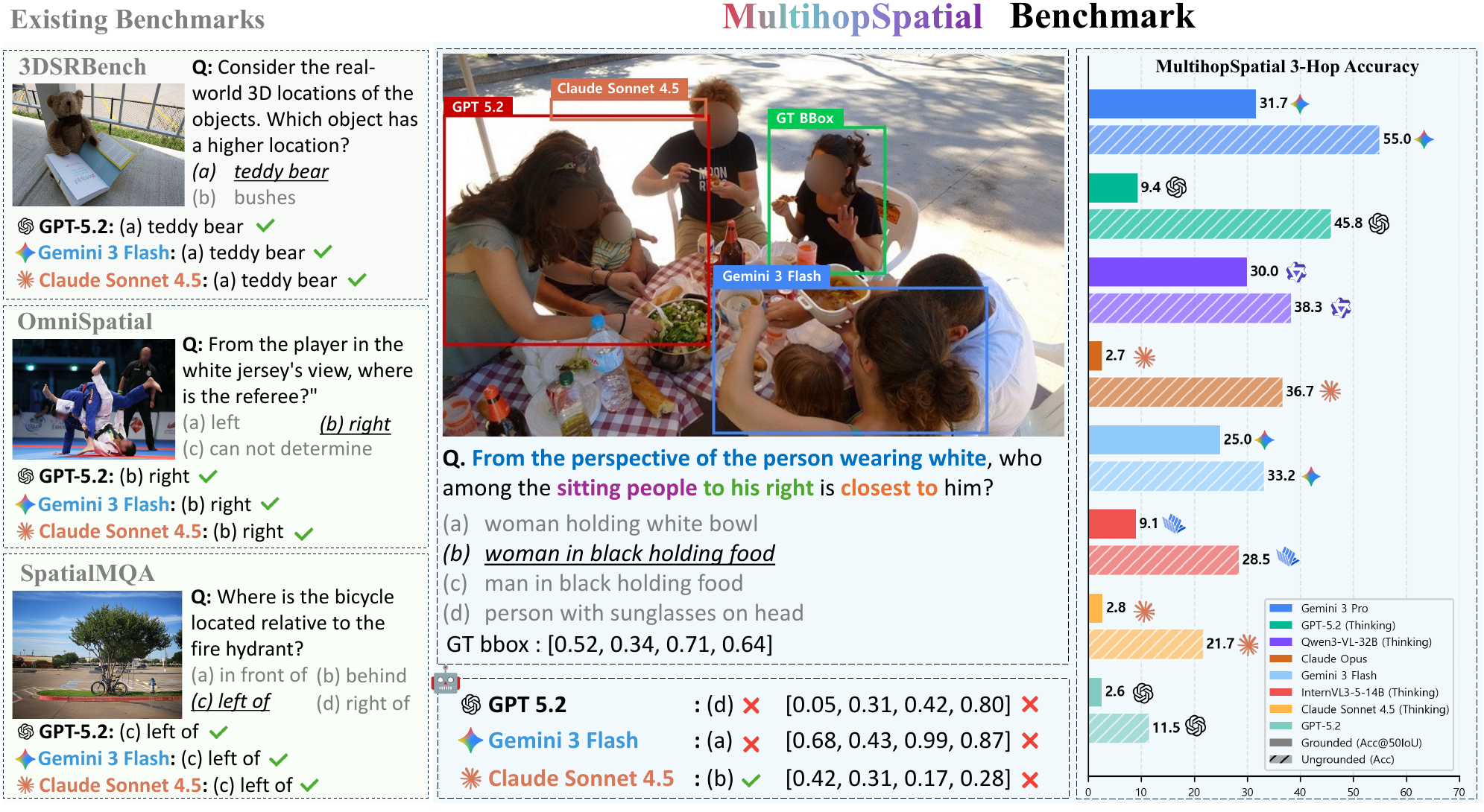}
        \vspace{-0.6cm}
    \caption{\textbf{Comparison of existing benchmarks~(single-hop) and MultihopSpatial benchmark.} In the question text, colored spans denote the queried reasoning components:
\textcolor{persp}{Perspective},
\textcolor{attri}{Attribute},
\textcolor{posit}{Position},
and \textcolor{relat}{Relation}.
    \vspace{-0.5cm}
    }\label{fig:teaser_2}
\end{figure}

VLM spatial reasoning benchmarks have rapidly evolved from elementary relations (SpatialVLM~\cite{spatialVLM}, BLINK~\cite{fu2024blink}) to diverse dimensions, including 3D properties (3DSRBench~\cite{3dsrbench}), video (VSI-Bench~\cite{vsibench}), scale (SpatialScore~\cite{spatialscore}), real-world complexity (SpatiaLab~\cite{spatialab}), fine-grained taxonomies (OmniSpatial~\cite{omnispatial25}), multi-image contexts (MMSI-Bench~\cite{yang2025mmsi}), and perspectives (SpatialMQA~\cite{spatialmqa}). However, they predominantly rely on single-hop queries, under-evaluating the compositional reasoning essential for real-world scenarios. MultihopSpatial addresses this by introducing 1- to 3-hop complex queries. Crucially, we advance beyond standard MCQs with a grounded metric requiring both correct answer selection and precise spatial localization, ensuring evaluations explicitly reflect the visual grounding vital for VLA deployment.
\section{MultihopSpatial}
\label{multihopSpatial}
\begin{figure}[t]
    \centering
        \includegraphics[width=\linewidth]{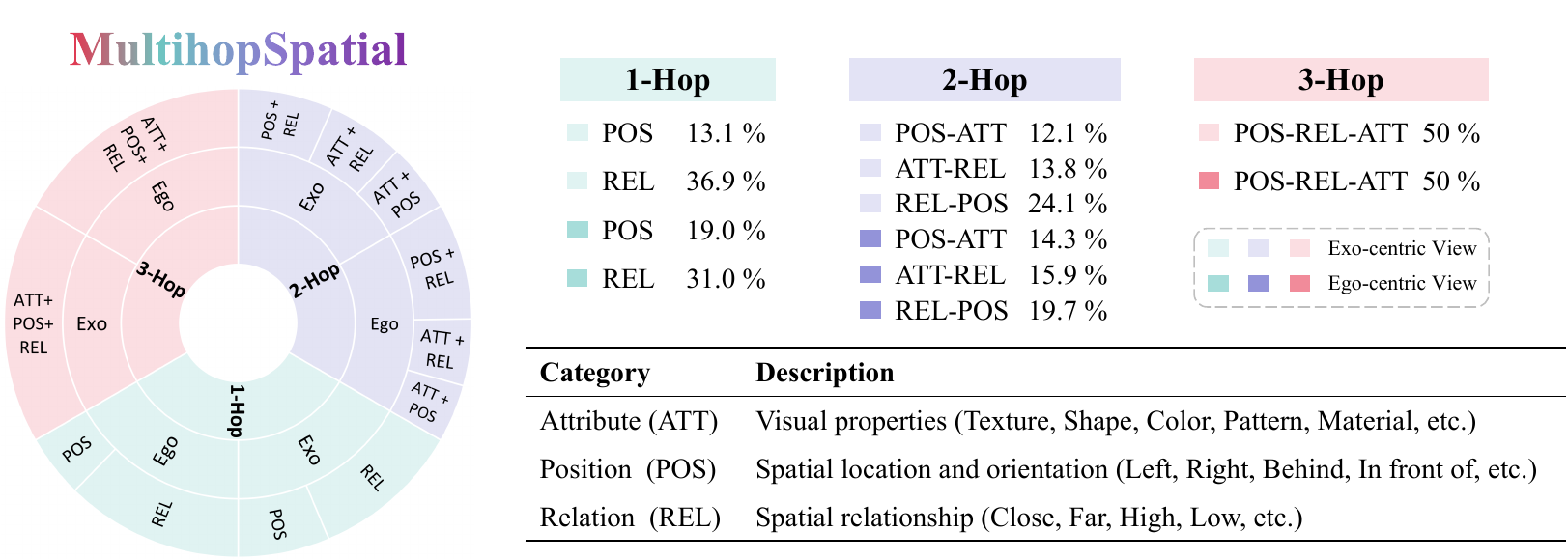}
        \vspace{-0.5cm}
    \caption{\textbf{Compositional structure and category definitions in MultihopSpatial.} }
    \vspace{-0.5cm}
    \label{fig:hop_statistic}
\end{figure}
\begin{figure}[t]
    \centering
        \includegraphics[width=\linewidth]{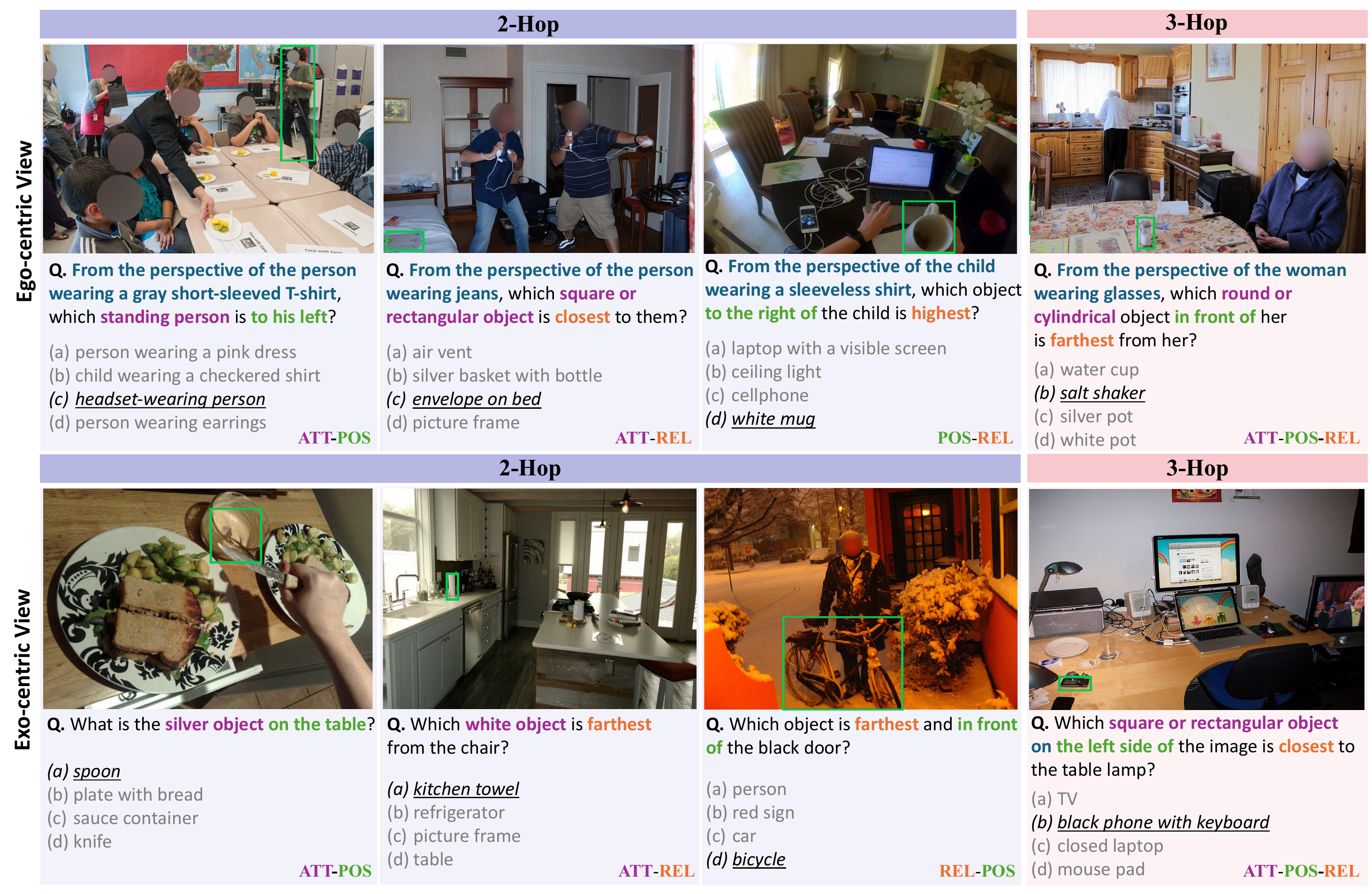}
    \vspace{-0.6cm}
    \caption{\textbf{Example MultihopSpatial questions for 2-hop and 3-hop reasoning under ego-centric (top) and exo-centric (bottom) views}. We omit the phrase ``\texttt{and provide the bounding box coordinate of the region related to your answer}'' for brevity.
    }
    \vspace{-0.5cm}
    \label{fig:hop_examples}
\end{figure}
\subsection{Overview}
We introduce \textbf{MultihopSpatial}, a 4.5K-sample benchmark of manually annotated VQA pairs focusing on multi-hop, compositional spatial queries common in real-world scenarios. Each example provides ground-truth bounding boxes to jointly evaluate reasoning and spatial localization, which is crucial for embodied/VLA settings (\eg, grasping and action). This grounded design improves interpretability and eliminates the evaluation blind spot of random guessing.

\subsection{Dataset Construction}
\vspace{2pt}
\noindent\textbf{Data Source, Annotation and Verification.}\quad
We curate 3,563 spatially complex images from COCO~\cite{COCO} and PACO-Ego4D~\cite{paco}, ensuring diverse coverage of everyday indoor/outdoor scenes and ego/exo perspectives.
Upon these images, we construct \textbf{4,500} multiple-choice questions (MCQs) for the proposed \textbf{MultihopSpatial benchmark}, perfectly balanced across 1- to 3-hop reasoning levels (1,500 per hop) and viewpoints (750 ego-centric and 750 exo-centric per hop).

To completely eliminate the reliability concerns and hallucinations inherent in AI-generated data, all QA pairs and bounding boxes were strictly annotated by ten trained human experts. Each sample underwent a rigorous multi-stage verification process with three rounds of independent cross-checking. Finally, three verifiers ensured that: (i) all option entities exist in the image, (ii) the bounding-box annotation precisely matches the referred target, and (iii) the labeled answer is correct and uniquely supported by the question. This strict protocol yielded a high inter-annotator agreement (Krippendorff's $\alpha = 0.90$), ensuring a high-quality and dependable benchmark for evaluating complex spatial reasoning.
Full annotation guidelines and verification procedures are detailed in~\cref{app:anno}.
Furthermore, we introduce \textbf{MultihopSpatial-Train}, an auxiliary large-scale training corpus of \textbf{6,791} grounded VQA samples designed to support the post-training of VLMs for spatial intelligence.

\vspace{2pt}
\noindent\textbf{Spatial Reasoning Categories.}\quad
We define three fundamental spatial reasoning categories: \textsc{Attribute} (\textsc{att}), \textsc{Position} (\textsc{pos}), and \textsc{Relation} (\textsc{rel}), detailed in ~\cref{fig:hop_statistic}. Our benchmark systematically composes these categories into multi-hop questions~(\eg, 2- and 3-hop) that demand sequential intermediate inferences, \eg, identifying candidates, restricting the search space, comparing relations, and resolving ambiguities. This compositional structure makes each intermediate output constrain the next step's search space, rather than treating all conditions as a single parallel conjunctive filter. Such compositional reasoning is essential for robust real-world scene understanding and embodied interaction. By increasing the hop count, we represent longer reasoning chains and escalating difficulty, enabling fine-grained diagnosis of model performance across different levels of reasoning complexity.

\noindent\textbf{1-Hop.}\quad
Single-step questions targeting one spatial category (\textsc{pos} or \textsc{rel}). We exclude \textsc{att} as a standalone category, since attributes are primarily perceptual unless composed with spatial metrics. While 1-hop spatial reasoning has been explored in prior works~\cite{3dsrbench,spatialmqa,omnispatial25,spatialab}, we deliberately include it to serve as a controlled baseline for depth-wise comparisons against multi-hop compositions.

\vspace{2pt}
\noindent\textbf{2-Hop.}\quad
Questions combining two categories (\textsc{att+pos}, \textsc{att+rel}, or \textsc{pos+rel}). As shown in ~\cref{fig:hop_examples}, these queries typically follow a two-stage structure: (i) restricting the candidate set using one category, and (ii) identifying the target by applying the other. We define this as 2-hop'' because both constraints must be jointly satisfied, regardless of the specific inference order.

\vspace{2pt}
\noindent\textbf{3-Hop.}\quad
Questions incorporating all three categories (\textsc{att+pos+rel}) in a single query. An \textsc{att} cue narrows candidates, \textsc{pos} further restricts the spatially valid subset, and \textsc{rel} selects the final target within that subset (\eg, selecting the rightmost object that is farthest/closest). This structure (~\cref{fig:hop_examples}) mirrors how humans refer to objects in cluttered scenes, testing the disambiguation needed for embodied task execution.

\section{Reinforcement Learning on MultihopSpatial-Train}
\label{sec:rl}

Beyond serving as a static evaluation benchmark, we investigate the utility of MultihopSpatial as a training corpus. To validate whether it can effectively enhance the multi-hop spatial reasoning capabilities of VLMs, we post-train a base VLM~(\eg, Qwen3-VL-4B-Instruct~\cite{qwen3vl}) via reinforcement learning. We adopt the Reinforcement Learning with Verifiable Rewards (RLVR) paradigm~\cite{deepseek-r1, deepseekmath}, employing deterministic, rule-based reward functions to provide unambiguous correctness signals. Our dataset is naturally suited for RLVR, as each sample provides a unique multiple-choice answer and a ground-truth bounding box for binary verification and continuous localization evaluation, respectively.

To optimize the VLM, we employ Group Relative Policy Optimization (GRPO), an efficient RL algorithm that eliminates the need for a critic model. For a given input, GRPO samples a group of responses and optimizes the policy using advantage scores obtained by normalizing sequence-level rewards within the group. This optimization is driven by our composite reward function, which jointly evaluates format adherence, answer correctness, and spatial localization accuracy.

The model is prompted to output a specific format: ``\texttt{Answer: (X)}'' followed by ``\texttt{Bounding Box: [x1, y1, x2, y2]}'', where \texttt{X} $\in \{a,b,c,d\}$ and coordinates are normalized to a $[0, 1000]$ scale~\cite{gemini, qwen3vl}. The total reward $R$ for a generated sequence is simply defined as the sum of three components:
\begin{equation}
R = R_{\text{format}} + \alpha \cdot R_{\text{mcq}} + \beta \cdot R_{\text{bbox}},
\label{eq:total_reward}
\end{equation}
where we set weighting coefficients $\alpha = \beta = 1$.

\noindent \textbf{Format Reward ($R_{\text{format}}$).} A binary reward assigning 1 if the output strictly follows the expected parsing format, and 0 otherwise.

\noindent \textbf{MCQ Reward ($R_{\text{mcq}}$).} A discrete correctness signal where $R_{\text{mcq}} = 1$ if the parsed choice matches the ground-truth label, and 0 otherwise. If parsing fails, it defaults to 0.

\noindent \textbf{Bounding Box Reward ($R_{\text{bbox}}$).} To evaluate spatial grounding, we calculate the Generalized Intersection over Union (GIoU)~\cite{giou} between the predicted and ground-truth bounding boxes. To provide a dense, positively-scaled training signal, we normalize the GIoU score ($\in [-1, 1]$) into a $[0, 1]$ range:
\begin{equation}
R_{\text{bbox}} = \frac{\mathrm{GIoU}(\hat{B}, B^{*}) + 1}{2}.
\label{eq:bbox_reward}
\end{equation}
A perfect overlap yields $R_{\text{bbox}} = 1$, whereas completely disjoint boxes receive $R_{\text{bbox}} < 0.5$. Unparsable bounding boxes default to 0.

By aggregating a discrete reasoning signal ($R_{\text{mcq}}$) with a continuous grounding signal ($R_{\text{bbox}}$), our composite reward design encourages the VLM to simultaneously enhance its compositional spatial reasoning and precise visual localization capabilities. Rather than optimizing solely for text-based answer correctness, this joint formulation ensures that the model's reasoning is intrinsically tied to actual visual grounding. Further training details are described in~\cref{app:train}.
We emphasize that our RL-based post-training serves as a deliberately \textbf{minimal baseline} to isolate the impact of our training data, excluding the advanced techniques used by specialized spatial reasoning models~\cite{remyxai2025spacethinker,ma2025spatialreasoner}; the reported gains thus represent a lower bound rather than a fully optimized result.

\section{Experiments}

\subsection{Experiment Setup}
We benchmark 37 VLMs spanning five categories on MultihopSpatial:

\noindent
(i)~\textbf{Proprietary instant models~(3)}: Claude-Opus-4.5~\cite{anthropic2025opus45},Claude--Sonnet-4.5~\cite{anthropic2025sonnet45}, and GPT-5.2~\cite{openai2025gpt52}.

\noindent
(ii)~\textbf{Proprietary reasoning models~(5)}: Gemini-3-Pro\footnote{\scriptsize \href{https://ai.google.dev/gemini-api/docs/thinking}{Gemini-3-Pro\&-Flash} operate with thinking mode enabled by default, and do not provide an option to fully disable it. We therefore classify them as reasoning models.}~\&-Flash, GPT-5.2-Thinking~(xhigh), Claude-Opus-4.5-Thinking, and Claude-Sonnet-4.5-Thinking.

\noindent
(iii)~\textbf{Open-weight instant models~(13)}: Qwen3-VL-Instruct~\cite{qwen3vl}~(4B,8B,32B, and 235B-A22B), InternVL-3.5~\cite{wang2025internvl3}~(4B,8B,14B,38B), GLM-4.6V~\cite{vteam2025glm}, Gemma-3-IT~\cite{kamath2025gemma} (4B,12B and 27B), and Molmo2-8B~\cite{clark2026molmo2}.

\noindent
(iv)~\textbf{Open-weight reasoning models~(9)}: Qwen3-VL-Thinking~(4B,8B,32B, and 235B-A22B), InternVL-3.5 with thinking mode~(4B,8B,14B,38B), and GLM-4.6V, each with thinking mode enabled.

\noindent
(v)~\textbf{Spatial reasoning models~(7)}: SenseNova-SI-1.3-InternVL3-8B~\cite{sensenova}, Cosmos-Reason2-8B~\cite{nvidia2024cosmosreason2}, VST-7B-RL~\cite{vst}, SpaceQwen3-VL-2B~\cite{spatialVLM}, SpaceOm~\cite{remyxai2024spaceom}, SpatialReasoner~\cite{ma2025spatialreasoner}, and SpaceThinker2.5VL-3B~\cite{remyxai2025spacethinker}. Reasoning variants are obtained by enabling the thinking mode of the corresponding instant model, using the \texttt{/think} tag or equivalent API option. All models are prompted with the same instruction template and evaluated under identical conditions. For each question, models are required to provide both a multiple-choice answer and a bounding box prediction for the target object (See the system prompt in~\cref{app:prompt}.)

\noindent
\textbf{Training setup.}
We train Qwen3-VL-4B-Instruct~\cite{qwen3vl} as the base policy model using GRPO~\cite{deepseekmath} with LoRA~\cite{lora} applied to the LLM backbone (10 epochs, learning rate $5 \times 10^{-5}$, batch size 128). For VLA evaluation, we integrate our model into VLM4VLA~\cite{zhang2026vlmvla} and train on the CALVIN ABC$\to$D~\cite{mees2022calvin} and Libero~\cite{liu2023libero}, following VLM4VLA hyperparameters. Please see details in~\cref{app:train}.
\definecolor{bestgreen}{HTML}{D1FAE5}

\begin{table*}[t]
\centering
\caption{\textbf{Benchmark results across different hop counts and Ego/Exo perspectives}. \colorbox{bestgreen}{Green} cells indicate the best performance within each model group.}
\vspace{-0.4cm}
\label{tab:main_results}
\resizebox{\textwidth}{!}{%
\begin{tabular}{l|ccc|cc|cc|cc|cc|cc|cc}
\toprule
\multirow{2}{*}{\textbf{Model}} & \multicolumn{3}{c|}{\textbf{Overall}} & \multicolumn{2}{c|}{\textbf{3Hop-Ego}} & \multicolumn{2}{c|}{\textbf{3Hop-Exo}} & \multicolumn{2}{c|}{\textbf{2Hop-Ego}} & \multicolumn{2}{c|}{\textbf{2Hop-Exo}} & \multicolumn{2}{c|}{\textbf{1Hop-Ego}} & \multicolumn{2}{c}{\textbf{1Hop-Exo}} \\
& Acc. & Acc@50 & avg. IoU & Acc. & Acc@50 & Acc. & Acc@50 & Acc. & Acc@50 & Acc. & Acc@50 & Acc. & Acc@50 & Acc. & Acc@50 \\
\midrule
\multicolumn{16}{l}{\textit{\textbf{Proprietary Models -- Instant}}} \\
Claude-Opus-4.5~\cite{anthropic2025opus45} & \cellcolor{bestgreen}45.1 & \cellcolor{bestgreen}3.2 & \cellcolor{bestgreen}13.3 & \cellcolor{bestgreen}25.7 & \cellcolor{bestgreen}2.0 & \cellcolor{bestgreen}48.5 & 3.6 & \cellcolor{bestgreen}33.7 & \cellcolor{bestgreen}2.0 & \cellcolor{bestgreen}58.1 & 4.8 & \cellcolor{bestgreen}43.2 & \cellcolor{bestgreen}3.5 & \cellcolor{bestgreen}61.1 & \cellcolor{bestgreen}3.1 \\
Claude-Sonnet-4.5~\cite{anthropic2025sonnet45} & 20.9 & 0.5 & 4.2 & 6.4 & 0.4 & 18.5 & 0.9 & 12.3 & 0.0 & 34.7 & 1.6 & 13.5 & 0.0 & 40.0 & 0.1 \\
GPT-5.2~\cite{openai2025gpt52} & 19.3 & 2.0 & 11.8 & 5.3 & 0.7 & 18.0 & \cellcolor{bestgreen}4.9 & 10.8 & 0.1 & 30.4 & \cellcolor{bestgreen}5.9 & 8.1 & 0.0 & 43.3 & 0.4 \\
\midrule
\multicolumn{16}{l}{\textit{\textbf{Open-weight Models -- Instant}}} \\
GLM-4.6V~\cite{vteam2025glm} & \cellcolor{bestgreen}43.2 & \cellcolor{bestgreen}35.2 & 69.5 & 15.9 & \cellcolor{bestgreen}12.3 & \cellcolor{bestgreen}46.7 & \cellcolor{bestgreen}39.3 & 22.7 & 18.4 & \cellcolor{bestgreen}61.6 & \cellcolor{bestgreen}53.2 & \cellcolor{bestgreen}32.4 & \cellcolor{bestgreen}24.3 & \cellcolor{bestgreen}80.1 & 63.7 \\
Molmo2-8B~\cite{clark2026molmo2} & 41.8 & 0.3 & 8.8 & 15.9 & 0.3 & 44.4 & 0.4 & 21.6 & 0.3 & 60.4 & 0.1 & \cellcolor{bestgreen}32.4 & 0.4 & 76.4 & 0.3 \\
Qwen3-VL-235B-Instruct~\cite{qwen3vl} & 41.3 & 34.8 & \cellcolor{bestgreen}71.1 & 14.8 & \cellcolor{bestgreen}12.3 & 42.9 & 37.6 & 21.9 & \cellcolor{bestgreen}18.5 & 58.5 & 52.7 & 30.7 & 23.5 & 79.2 & \cellcolor{bestgreen}64.4 \\
Qwen3-VL-32B-Instruct~\cite{qwen3vl} & 40.9 & 33.4 & 69.6 & 13.9 & 9.7 & 43.9 & 36.4 & 21.9 & 17.2 & 59.2 & 53.1 & 30.4 & 22.3 & 76.4 & 61.6 \\
InternVL-3.5-38B~\cite{wang2025internvl3} & 40.8 & 9.7 & 28.7 & 17.5 & 3.2 & 44.5 & 12.3 & \cellcolor{bestgreen}24.3 & 4.5 & 56.8 & 24.5 & 31.9 & 3.3 & 69.6 & 10.4 \\
InternVL-3.5-14B~\cite{wang2025internvl3} & 39.7 & 7.9 & 26.2 & 17.1 & 2.5 & 44.5 & 10.1 & 22.0 & 2.8 & 56.1 & 14.4 & 30.4 & 3.3 & 68.0 & 14.0 \\
InternVL-3.5-4B~\cite{wang2025internvl3} & 39.7 & 10.3 & 30.0 & 15.9 & 3.1 & 45.2 & 13.5 & 23.9 & 4.5 & 54.9 & 18.7 & 31.9 & 3.6 & 66.3 & 18.4 \\
InternVL-3.5-8B~\cite{wang2025internvl3} & 38.3 & 9.4 & 30.0 & 14.7 & 2.9 & 41.1 & 10.7 & 23.3 & 4.0 & 50.1 & 17.2 & 30.5 & 4.5 & 70.0 & 17.3 \\
Qwen3-VL-8B-Instruct~\cite{qwen3vl} & 38.0 & 31.3 & 69.5 & 12.3 & 8.8 & 42.1 & 36.1 & 18.8 & 15.2 & 55.3 & 49.1 & 26.7 & 20.8 & 72.8 & 58.0 \\
Qwen3-VL-4B-Instruct~\cite{qwen3vl} & 37.8 & 31.0 & 69.9 & 15.5 & 10.9 & 40.1 & 33.7 & 20.9 & 16.0 & 53.5 & 46.9 & 26.7 & 21.2 & 70.3 & 57.2 \\
Gemma-3-IT-27B~\cite{kamath2025gemma} & 33.1 & 0.4 & 5.4 & 18.1 & 0.1 & 30.8 & 0.5 & 22.0 & 0.1 & 45.7 & 1.1 & 28.1 & 0.3 & 53.9 & 0.3 \\
Gemma-3-IT-12B~\cite{kamath2025gemma} & 29.8 & 0.4 & 5.9 & 16.9 & 0.3 & 31.6 & 0.5 & 22.1 & 0.4 & 40.8 & 0.5 & 22.9 & 0.1 & 44.5 & 0.5 \\
Gemma-3-IT-4B~\cite{kamath2025gemma} & 28.4 & 0.2 & 3.0 & \cellcolor{bestgreen}22.1 & 0.0 & 27.2 & 0.3 & 22.5 & 0.1 & 36.8 & 0.3 & 27.1 & 0.3 & 34.7 & 0.1 \\
\midrule
\multicolumn{16}{l}{\textit{\textbf{Proprietary Models -- Reasonin}g}} \\
Gemini-3-Pro~\cite{gemini} & \cellcolor{bestgreen}64.7 & \cellcolor{bestgreen}40.6 & 55.0 & \cellcolor{bestgreen}39.7 & \cellcolor{bestgreen}18.8 & \cellcolor{bestgreen}71.1 & 45.3 & 36.8 & 20.5 & \cellcolor{bestgreen}81.2 & 55.5 & \cellcolor{bestgreen}71.1 & \cellcolor{bestgreen}41.1 & \cellcolor{bestgreen}88.4 & \cellcolor{bestgreen}62.3 \\
GPT-5.2-Thinking~\cite{openai2025gpt52} & 57.9 & 11.5 & 29.0 & 36.1 & 8.5 & 55.7 & 10.4 & \cellcolor{bestgreen}49.7 & 7.6 & 63.6 & 18.0 & 65.6 & 12.5 & 76.4 & 11.7 \\
Gemini-3-Flash~\cite{gemini} & 57.2 & 40.2 & \cellcolor{bestgreen}61.2 & 6.9 & 4.3 & 61.2 & \cellcolor{bestgreen}46.9 & 42.3 & \cellcolor{bestgreen}25.3 & 80.0 & \cellcolor{bestgreen}63.7 & 66.0 & 38.9 & 86.8 & 62.1 \\
Claude-Opus-4.5-Thinking~\cite{anthropic2025opus45} & 47.0 & 4.7 & 16.7 & 25.5 & 3.5 & 49.7 & 4.7 & 35.2 & 3.1 & 60.0 & 8.8 & 45.1 & 5.2 & 66.5 & 2.9 \\
Claude-Sonnet-4.5-Thinking~\cite{anthropic2025sonnet45} & 32.2 & 4.3 & 19.2 & 14.7 & 1.9 & 29.9 & 3.6 & 22.1 & 2.3 & 45.7 & 8.1 & 31.3 & 3.9 & 49.3 & 6.1 \\
\midrule
\multicolumn{16}{l}{\textit{\textbf{Open-weight Models -- Reasoning}}} \\
Qwen3-VL-32B-Thinking~\cite{qwen3vl} & \cellcolor{bestgreen}46.8 & \cellcolor{bestgreen}37.4 & 67.2 & 19.2 & \cellcolor{bestgreen}12.9 & \cellcolor{bestgreen}57.5 & \cellcolor{bestgreen}47.1 & 24.3 & 18.1 & \cellcolor{bestgreen}70.1 & \cellcolor{bestgreen}60.0 & 30.4 & 23.1 & \cellcolor{bestgreen}79.6 & \cellcolor{bestgreen}63.2 \\
Qwen3-VL-235B-Thinking~\cite{qwen3vl} & 45.1 & 36.3 & 67.8 & 17.6 & 12.7 & 51.2 & 42.3 & \cellcolor{bestgreen}24.8 & \cellcolor{bestgreen}19.3 & 67.6 & 58.7 & 31.2 & 22.8 & 78.1 & 61.9 \\
Qwen3-VL-4B-Thinking~\cite{qwen3vl} & 42.6 & 28.7 & 58.3 & 20.4 & 9.2 & 48.1 & 34.8 & 22.3 & 12.0 & 63.9 & 50.7 & 30.8 & 16.3 & 70.4 & 49.2 \\
InternVL-3.5-38B-Thinking~\cite{wang2025internvl3} & 42.1 & 27.4 & 57.0 & 19.5 & 10.9 & 43.3 & 32.5 & 24.7 & 15.3 & 56.8 & 39.2 & \cellcolor{bestgreen}34.8 & 20.7 & 73.6 & 45.9 \\
GLM-4.6V-Thinking~\cite{vteam2025glm} & 42.0 & 34.7 & \cellcolor{bestgreen}70.1 & 14.1 & 10.7 & 46.3 & 40.0 & 19.5 & 15.3 & 63.1 & 56.1 & 31.2 & \cellcolor{bestgreen}23.3 & 77.6 & 62.7 \\
Qwen3-VL-8B-Thinking~\cite{qwen3vl} & 41.7 & 29.5 & 60.1 & 18.5 & 9.2 & 47.9 & 36.4 & 21.3 & 11.9 & 63.3 & 51.6 & 28.5 & 16.5 & 70.5 & 51.2 \\
InternVL-3.5-8B-Thinking~\cite{wang2025internvl3} & 40.6 & 5.2 & 22.1 & \cellcolor{bestgreen}20.5 & 1.3 & 39.9 & 6.7 & 24.1 & 1.9 & 57.3 & 10.4 & 30.9 & 3.5 & 70.5 & 7.6 \\
InternVL-3.5-4B-Thinking~\cite{wang2025internvl3} & 40.6 & 4.7 & 21.4 & 18.3 & 1.2 & 41.2 & 4.8 & 24.4 & 1.7 & 58.5 & 9.6 & 32.8 & 1.5 & 68.1 & 9.3 \\
InternVL-3.5-14B-Thinking~\cite{wang2025internvl3} & 38.2 & 11.1 & 34.7 & 14.1 & 4.7 & 42.8 & 13.6 & 21.1 & 5.6 & 55.3 & 20.5 & 27.6 & 6.4 & 68.0 & 15.9 \\
\midrule
\multicolumn{16}{l}{\textit{\textbf{Specialized Spatial Reasoning Models}}} \\
SenseNova-InternVL3-8B~\cite{sensenova} & \cellcolor{bestgreen}42.3 & 17.3 & 38.8 & \cellcolor{bestgreen}20.4 & 9.1 & \cellcolor{bestgreen}45.2 & 19.7 & \cellcolor{bestgreen}25.2 & 9.2 & \cellcolor{bestgreen}55.5 & 27.2 & \cellcolor{bestgreen}34.5 & 11.2 & \cellcolor{bestgreen}73.2 & 27.2 \\
Cosmos-Reason2-8B~\cite{nvidia2024cosmosreason2} & 37.8 & \cellcolor{bestgreen}27.9 & 61.4 & 15.2 & \cellcolor{bestgreen}10.5 & 40.7 & \cellcolor{bestgreen}31.9 & 19.5 & \cellcolor{bestgreen}13.5 & 54.5 & \cellcolor{bestgreen}43.5 & 26.5 & \cellcolor{bestgreen}17.1 & 70.1 & \cellcolor{bestgreen}51.1 \\
VST-7B-RL~\cite{vst} & 36.0 & 16.2 & 42.4 & 16.7 & 8.5 & 34.1 & 16.1 & 23.9 & 8.9 & 48.1 & 28.3 & 24.5 & 9.6 & 68.8 & 26.0 \\
SpaceQwen3-VL-2B~\cite{spatialVLM} & 33.6 & 10.1 & 31.5 & 18.5 & 4.0 & 32.5 & 9.9 & 22.8 & 4.0 & 47.2 & 22.9 & 26.1 & 4.4 & 54.3 & 15.2 \\
SpaceOm~\cite{remyxai2024spaceom} & 32.3 & 22.7 & \cellcolor{bestgreen}61.8 & 15.3 & 9.5 & 37.9 & 26.4 & 19.6 & 12.0 & 47.9 & 37.2 & 20.5 & 13.2 & 52.8 & 37.9 \\
SpatialReasoner~\cite{ma2025spatialreasoner} & 31.7 & 8.7 & 29.9 & 18.0 & 6.8 & 34.0 & 10.3 & 19.6 & 4.3 & 46.0 & 13.7 & 21.3 & 5.7 & 51.5 & 11.2 \\
SpaceThinker-3B~\cite{remyxai2025spacethinker} & 31.1 & 14.4 & 45.2 & 15.9 & 6.9 & 36.3 & 18.8 & 19.2 & 5.2 & 44.5 & 27.3 & 20.8 & 6.5 & 50.0 & 21.5 \\
\bottomrule
\end{tabular}%
}
\vspace{-0.5cm}
\end{table*}

\subsection{Evaluation Metrics}
\label{sec:eval_metrics}

We employ three complementary metrics to jointly evaluate reasoning correctness and spatial grounding:

\noindent
\textbf{MCQ Accuracy.}
Measures the percentage of correct multiple-choice predictions ($\hat{y} = y^{*}$). While standard, it does not verify spatial localization.

\noindent
\textbf{Acc@50IoU.}
Our primary grounded metric requires correct answer selection and precise localization. 
A prediction is correct only if $\hat{y} = y^{*}$ and $\mathrm{IoU}(\hat{B}, B^{*}) \geq 0.5$. This filters out ungrounded predictions, ensuring genuine localization.

\noindent
\textbf{Avg.\ IoU.}
Computed exclusively over MCQ-correct samples, this metric isolates grounding capability from reasoning errors, evaluating how precisely a model localizes the target once correctly identified.

\vspace{-1em}
\subsection{Main Results}
\begin{figure*}[t]
    \centering
        \includegraphics[width=\linewidth]{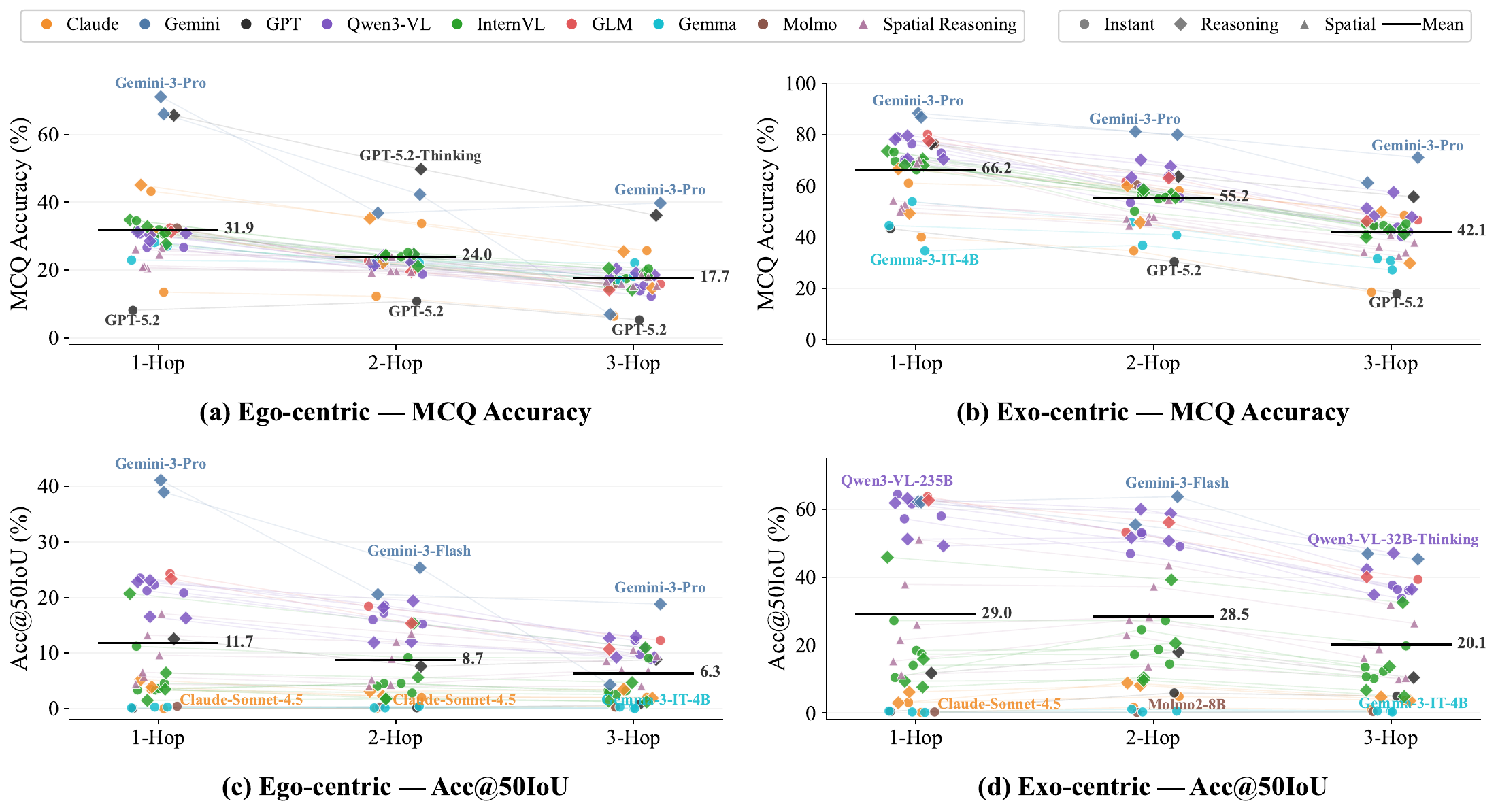}
    \vspace{-0.6cm}
    \caption{\textbf{All model performance across hop count and ego/exo perspectives.}}
    \vspace{-0.6cm}
    \label{fig:scatter}
\end{figure*}
\textbf{Overall Performance Highlights.} As shown in~\cref{tab:main_results}, Gemini-3-Pro achieves the highest MCQ accuracy (64.7\%) and Acc@50IoU (40.6\%), while Qwen3-VL-32B-Thinking leads the open-weight category. Notably, top performance in answer selection does not guarantee precise localization (\eg, Qwen3-VL-235B leads in average IoU). This discrepancy demonstrates that compositional reasoning and grounding capabilities remain largely decoupled in current VLMs.

\noindent
\textbf{Metric-dependent Rankings.} This decoupling causes drastic rank reversals between metrics. Models like Claude-Opus-4.5 and Molmo2-8B rank high in MCQ but plummet in Acc@50IoU (\eg, Claude drops from 7th to 29th), indicating shortcut-based answers without genuine localization. Conversely, the smaller Qwen3-VL-4B rises from 25th to 10th due to its robust grounding. These shifts prove that evaluating MCQ alone is highly misleading, making Acc@50IoU essential for verifying true spatial understanding.

\noindent
\textbf{Benchmark Difficulty.} With the best model peaking at just 40.6\% Acc@50IoU, MultihopSpatial remains far from saturated. The difficulty peaks under 3-hop ego-centric conditions, where only 3 of 37 models exceed the 25\% random MCQ baseline, and just 9 surpass 10\% Acc@50IoU. Strikingly, even advanced reasoning models like GPT-5.2-Thinking (8.5\%) and Claude-Sonnet-4.5-Thinking (1.9\%) fail drastically here, confirming our benchmark rigorously evaluates both compositional reasoning and spatial grounding capabilities of current VLMs.

\subsection{In-depth Analysis}
In this section, we present a comprehensive analysis of 37 VLMs on the proposed MultihopSpatial benchmark and derive eight key insights.

\begin{figure*}[t]
    \centering
    \begin{minipage}[t]{0.41\textwidth}
        \centering
        \includegraphics[width=\linewidth]{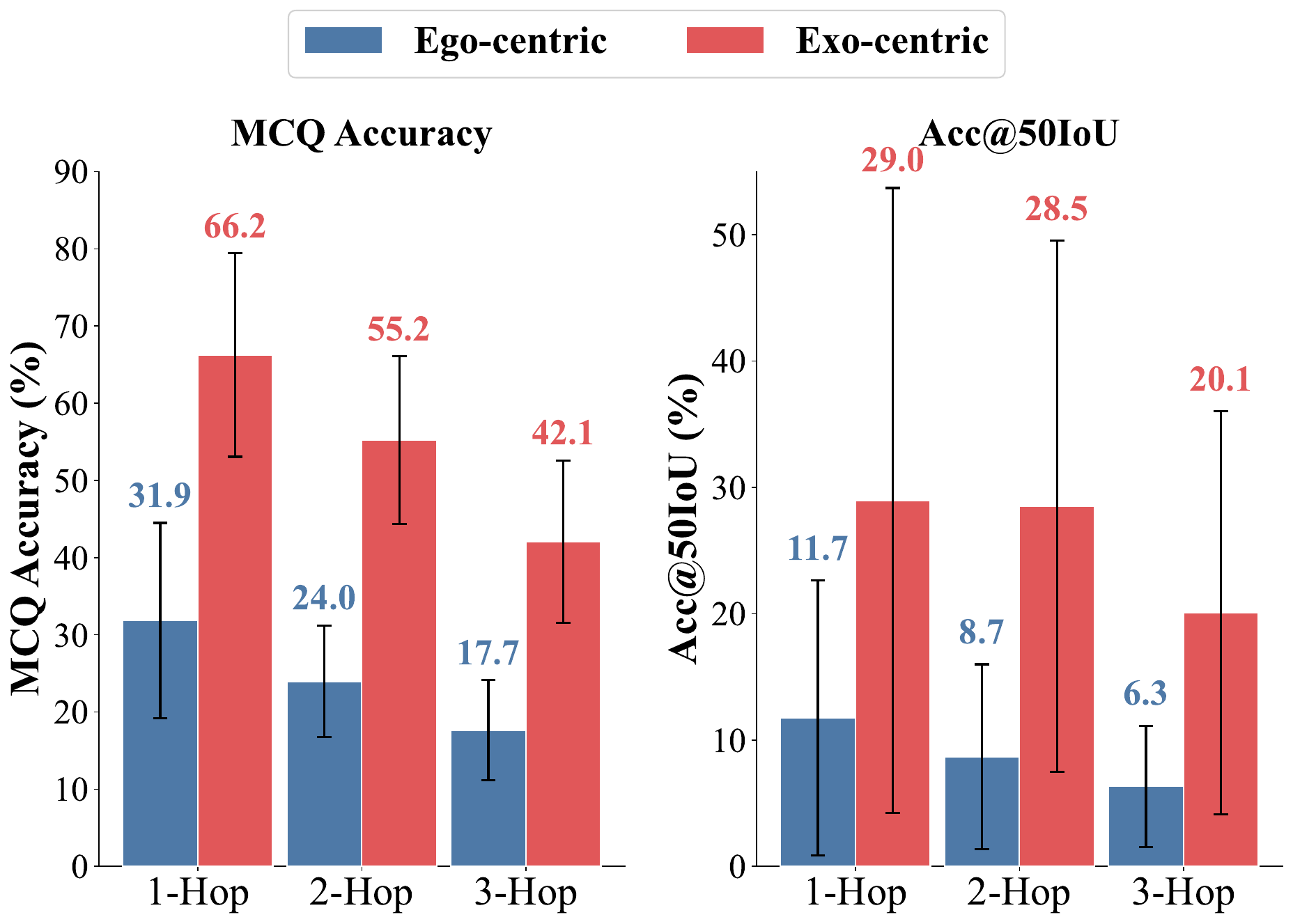}
        \vspace{-0.5cm}
        \caption{\textbf{Average Performance by Hop Count.}}
        \vspace{-0.5cm}
        \label{fig:a1_hop}
    \end{minipage}
    \hfill
    \begin{minipage}[t]{0.58\textwidth}
        \centering
        \includegraphics[width=\linewidth]{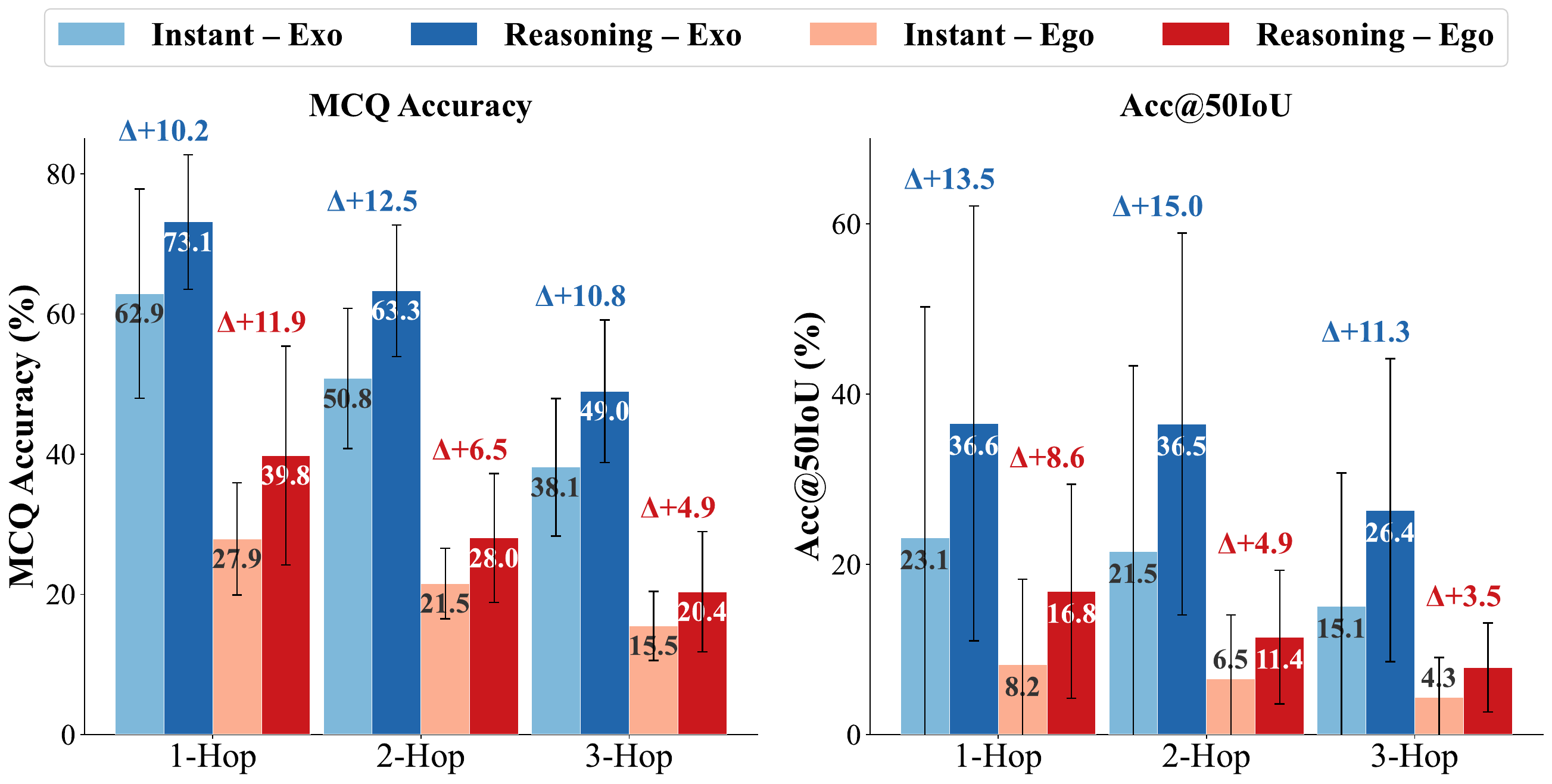}
        \vspace{-0.5cm}
        \caption{\textbf{Reasoning vs.\ Instant Models.}}
        \vspace{-0.5cm}
        \label{fig:a2_reason}
    \end{minipage}
\end{figure*}

\noindent
\textbf{A1. Performance Degradation across Reasoning Hops.}
Evaluating the impact of reasoning complexity across all 37 models (\cref{fig:a1_hop}) reveals a consistent performance degradation as the number of hops increases, confirming that compositional spatial reasoning remains a fundamental challenge for current VLMs. Both MCQ accuracy and Acc@50IoU exhibit steep declines from 1-hop to 3-hop. Crucially, this degradation is exacerbated under ego-centric evaluation, which requires additional perspective transformation: going from 1-hop to 3-hop, the ego-centric view retains only 55.5\% of its MCQ accuracy (31.9→17.7) against 63.6\% for the exo-centric view (66.2→42.1), and 53.8\% against 69.3\% under Acc@50IoU. The ego/exo accuracy ratio therefore shrinks from 0.48 to 0.42, suggesting that perspective-taking compounds with multi-step reasoning in a multiplicative rather than additive manner.

\begin{figure}[t]
    \centering
    \begin{minipage}[t]{0.54\textwidth}
        \centering
        \includegraphics[width=\linewidth]{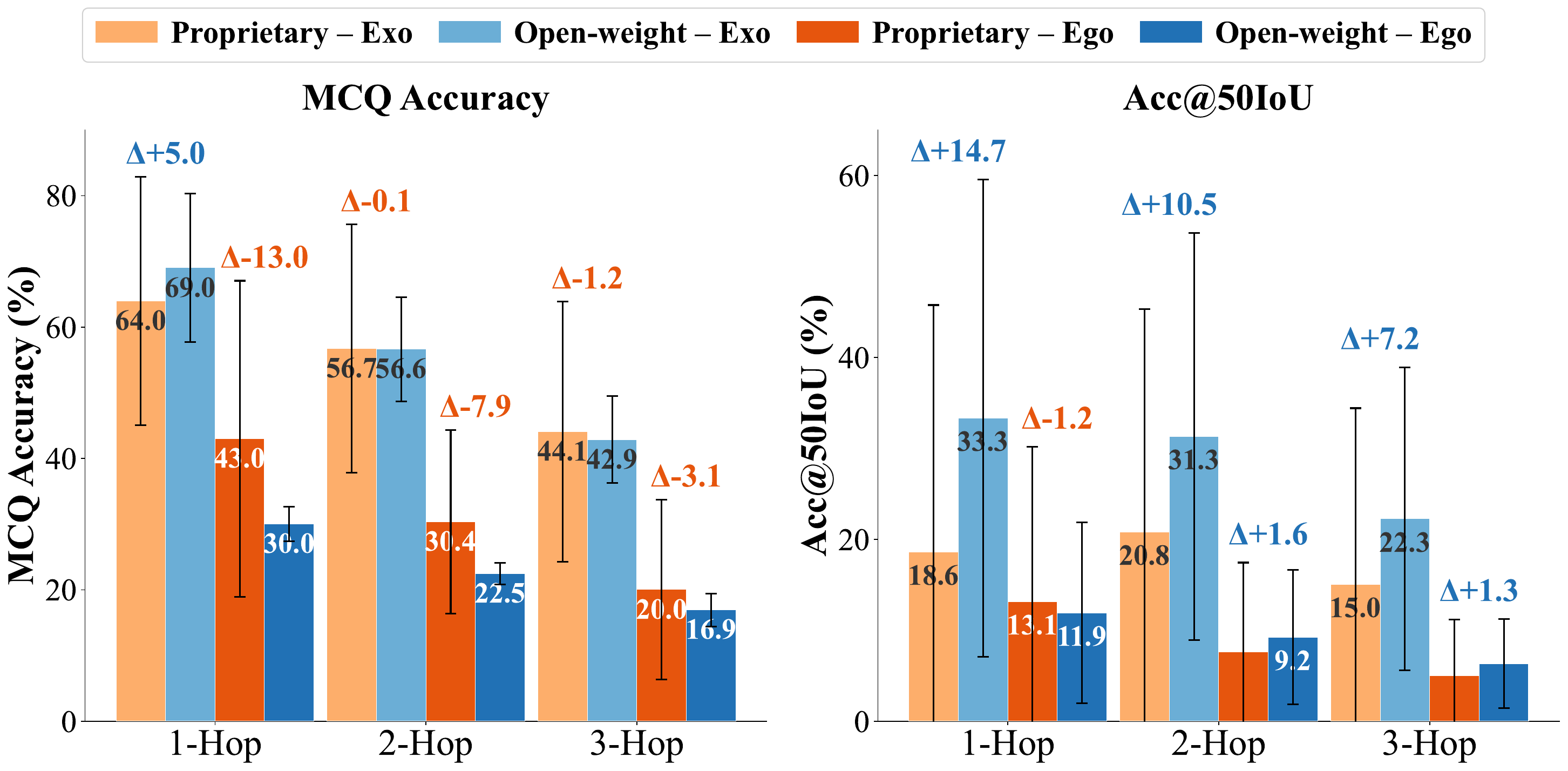}
        \vspace{-0.4cm}
        \captionof{figure}{\textbf{Open-weight vs.\ Proprietary models.}}
        \vspace{-0.4cm}
        \label{fig:open_vs_prop}
    \end{minipage}
    \vspace{-0.3cm}
    \hfill
    \begin{minipage}[t]{0.43\textwidth}
        \centering
        \includegraphics[width=\linewidth]{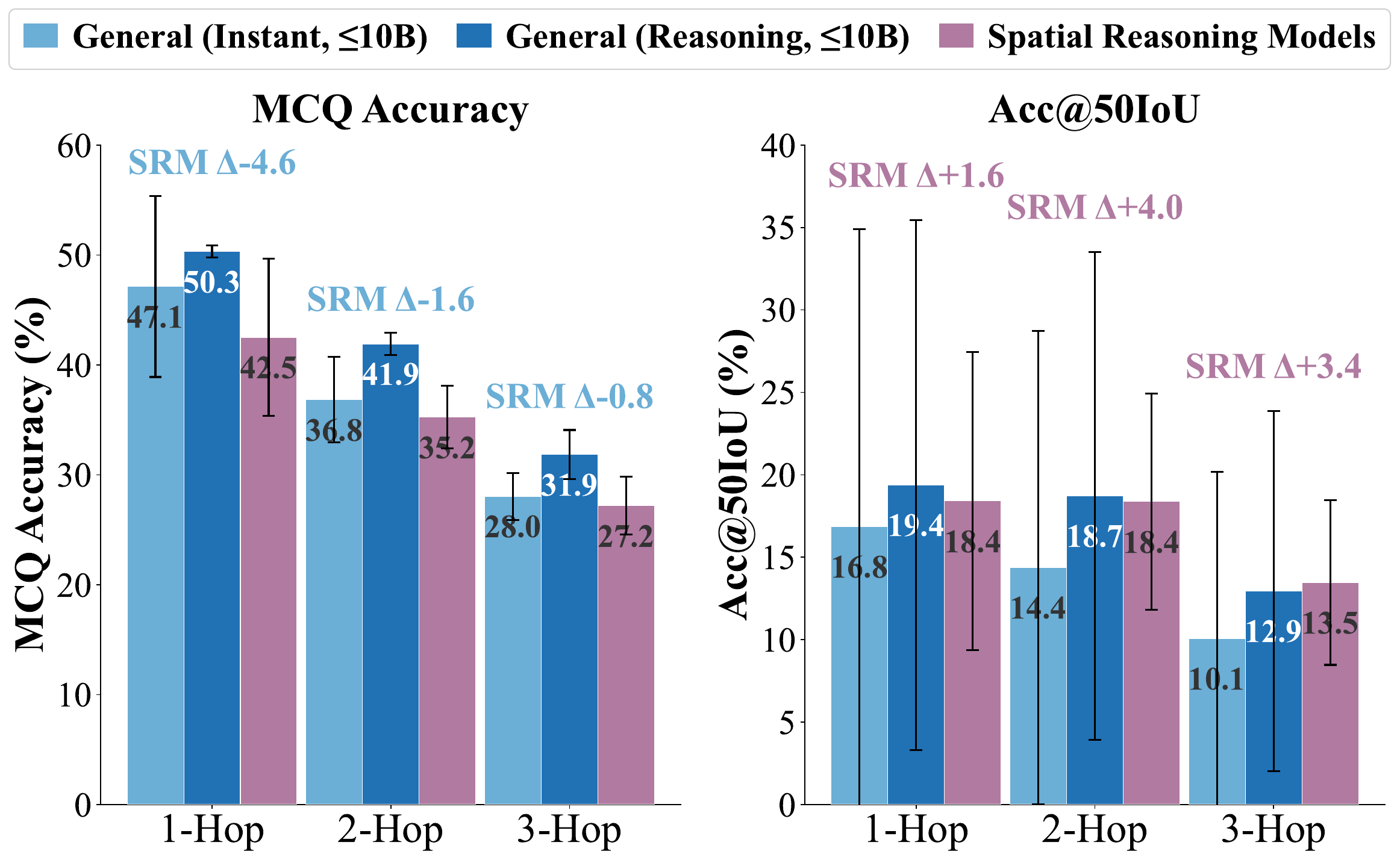}
        \vspace{-0.4cm}
        \captionof{figure}{\textbf{Spatial Reasoning Models vs.\ General Models ($\leq$10B).}}
        \vspace{-0.4cm}
        \label{fig:srm}
    \end{minipage}
\end{figure}

\noindent
\textbf{A2. Ego vs. Exo: Perspective-Taking as a Compounding Bottleneck.}
Beyond a simple performance drop (\cref{fig:a1_hop}), ego-centric evaluation fundamentally alters capability visibility. Under exo-centric conditions (\cref{fig:scatter}(d)), our Acc@50IoU metric clearly distinguishes grounding-capable models (\eg, Qwen3-VL) from those lacking native localization (\eg, InternVL-3.5) despite their similar MCQ accuracies. Conversely, ego-centric evaluation (\cref{fig:scatter}(c)) acts as an evaluation blind spot; it suppresses even strong grounding models to a 20–25\% floor, completely masking these capability gaps. This compression highlights the necessity of evaluating perspectives jointly and reinforces Acc@50IoU as an essential metric to expose disparities invisible to conventional MCQ evaluation.

\noindent
\textbf{A3. Instant vs. Reasoning: Diminishing Returns under Multi-Hop Pressure.}
While reasoning models consistently outperform instant models (\cref{fig:a2_reason}), this advantage diminishes under multi-hop pressure in the ego-centric setting: the MCQ gain shrinks from +11.9 pp at 1-hop to +4.9 pp at 3-hop (and from +8.6 to +3.5 pp under Acc@50IoU), whereas the exo-centric gain stays largely flat (+10.2 to +10.8 pp).Critically, even with extended thinking, reasoning models fall to roughly 20\% MCQ accuracy and below 10\% Acc@50IoU on 3-hop ego-centric tasks. This convergence toward the performance floor reveals that inference-time reasoning (\eg, chain-of-thought) yields diminishing returns as compositional steps accumulate. MultihopSpatial thus exposes a capability ceiling that test-time compute alone cannot resolve, confirming its value as a genuinely challenging benchmark even for state-of-the-art models.

\noindent
\textbf{A4. Open-weight vs. Proprietary Models.}
While proprietary models hold a slight MCQ advantage in ego-centric tasks (\cref{fig:open_vs_prop})—indicating stronger perspective-taking—open-weight models consistently dominate in Acc@50IoU. This reversal stems from a distinct asymmetry in visual grounding capabilities. Proprietary models exhibit high variance: while the Gemini-3 series demonstrates robust localization, others (\eg, GPT-5.2, Claude) lack native grounding, significantly dragging down the group average. Conversely, open-weight families (\eg, Qwen3-VL, GLM-4.6V) maintain consistently high grounding performance, forming a dense cluster at 60\%+ in exo-centric Acc@50IoU. These findings highlight that the open-weight ecosystem currently offers more reliable spatial grounding than its proprietary counterparts.

\begin{figure*}[t]
    \centering
    \includegraphics[width=\linewidth]{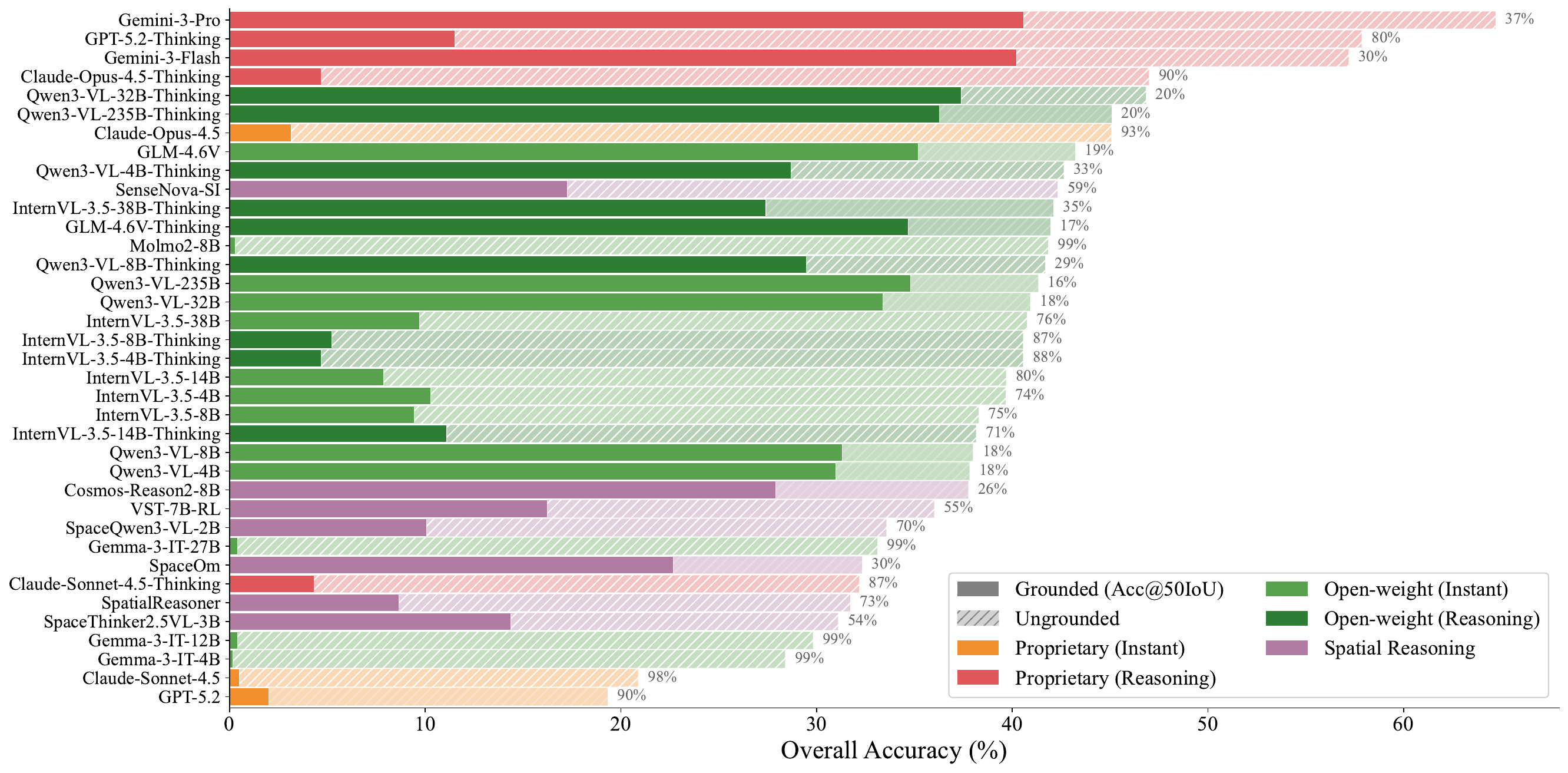}
    \vspace{-0.6cm}
    \caption{\textbf{Grounded~(Acc@50IoU) vs.\ Ungrounded accuracy per model.} Solid bars indicate grounded accuracy (Acc@50IoU) and hatched regions indicate ungrounded responses. Percentages denote the ungrounded ratio.}
    \label{fig:lucky_guess}
    \vspace{-0.3cm}
\end{figure*}
\begin{figure}[t]
    \centering
    \begin{minipage}[t]{0.62\textwidth}   
        \centering
        \includegraphics[width=\linewidth]{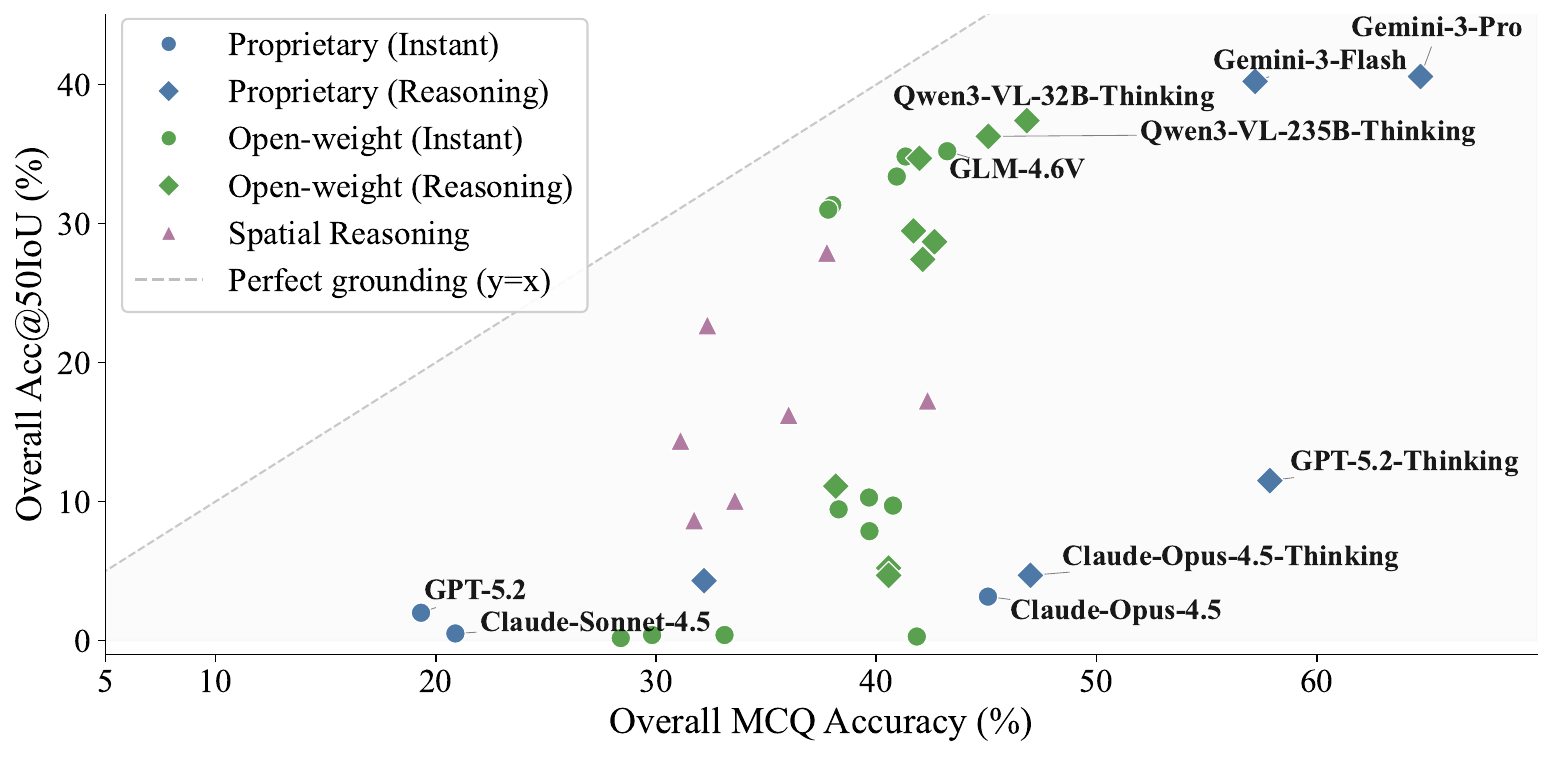}
        \vspace{-0.5cm}
        \captionof{figure}{\textbf{Grounding Gap.}}
        \label{fig:grounding_gap}
    \end{minipage}
    \hfill
    \begin{minipage}[t]{0.36\textwidth}   
        \centering
        \includegraphics[width=\linewidth]{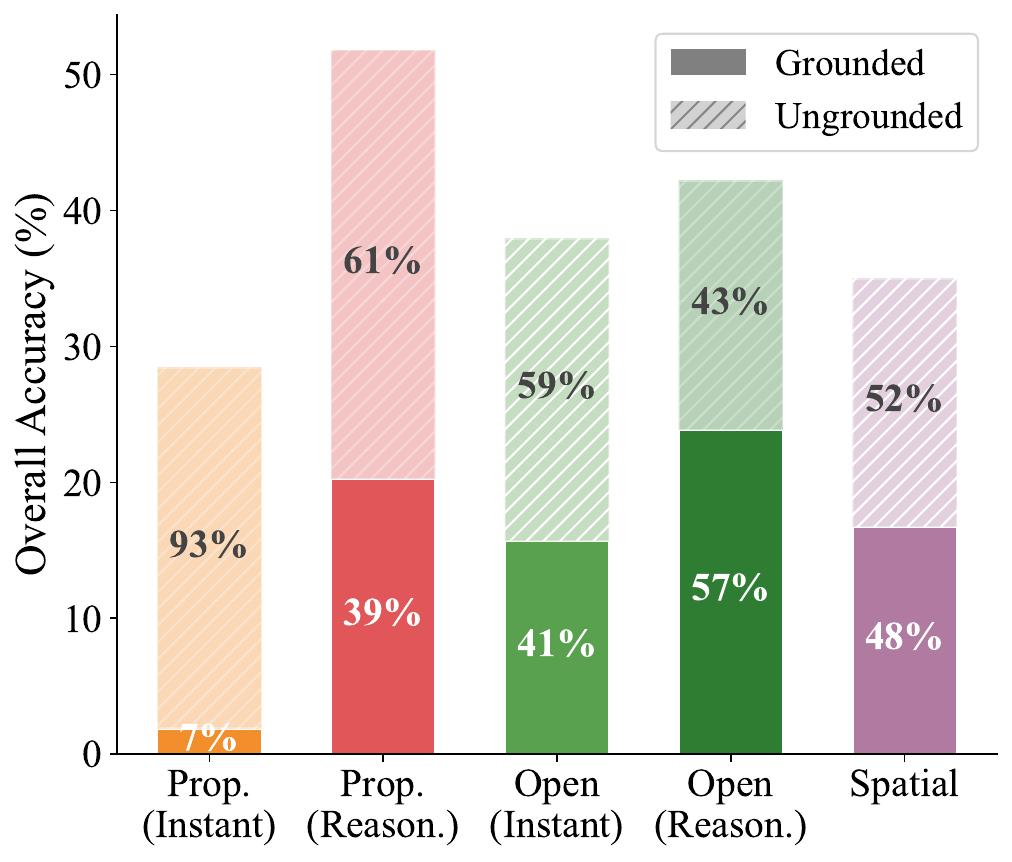}
        \vspace{-0.5cm}
        \captionof{figure}{\textbf{Ungrounded Ratio.}}
        \label{fig:lucky_cat}
    \end{minipage}
    \vspace{-0.5cm}
\end{figure}

\noindent
\textbf{A5. Generalist vs. Specialist Models.} We compare specialized spatial reasoning models (SRMs) with general-purpose VLMs of comparable scale ($\leq$10B) in \cref{fig:srm}. Despite being explicitly tailored for spatial reasoning, SRMs are consistently outperformed by generalists on multi-hop MCQ accuracy at every hop count, although the gap narrows monotonically as the reasoning chain deepens ($-4.6 \!\rightarrow\! -1.6 \!\rightarrow\! -0.8$ from 1- to 3-hop). This indicates that current spatial specialization does not transfer to the compositional, multi-hop reasoning that our benchmark targets. The ordering only reverses under Acc@50IoU, which counts a prediction as correct only when the target is additionally localized (IoU $\geq 50\%$): here SRMs surpass their instant generalist counterparts at every hop, and the margin widens with hop depth ($\Delta{+}1.6$, $+4.0$, $+3.4$). Since this metric further requires accurate localization, the reversal indicates that the advantage of SRMs is confined to grounding rather than reasoning---even with lower MCQ accuracy, they localize correctly answered targets more reliably. Generalists are thus stronger at answer selection while SRMs are stronger at localization, and these complementary strengths suggest that robust multi-hop spatial understanding requires the joint optimization of reasoning and grounding, which neither family currently attains.

\noindent
\textbf{A6. Grounding gap: Grounded vs. Ungrounded Accuracy.}
\cref{fig:grounding_gap} reveals a striking disconnect between answer selection and spatial localization: most models fall far below the $y=x$ line, indicating that high MCQ accuracy does not entail accurate grounding. Across all 37 models, the average ungrounded ratio reaches 59\%, meaning that more than half of correctly answered questions lack proper spatial localization (\cref{fig:lucky_guess}). This ratio varies dramatically by category (\cref{fig:lucky_cat}): proprietary instant models exhibit 93\% ungrounded ratio, while open-weight reasoning models achieve the lowest at 43\%, largely driven by Qwen3-VL and GLM families whose ungrounded ratios remain below 20\%. 
At the extreme, models such as Gemma-3-IT, Molmo2, and Claude-Sonnet-4.5 exceed 98\% ungrounded, effectively answering through shortcuts without any spatial understanding. 
These findings validate Acc@50IoU as an essential complement to MCQ evaluation: without it, models that rely entirely on linguistic shortcuts would be indistinguishable from those with genuine spatial understanding.

\begin{figure*}[t]
    \centering
    \includegraphics[width=\linewidth]{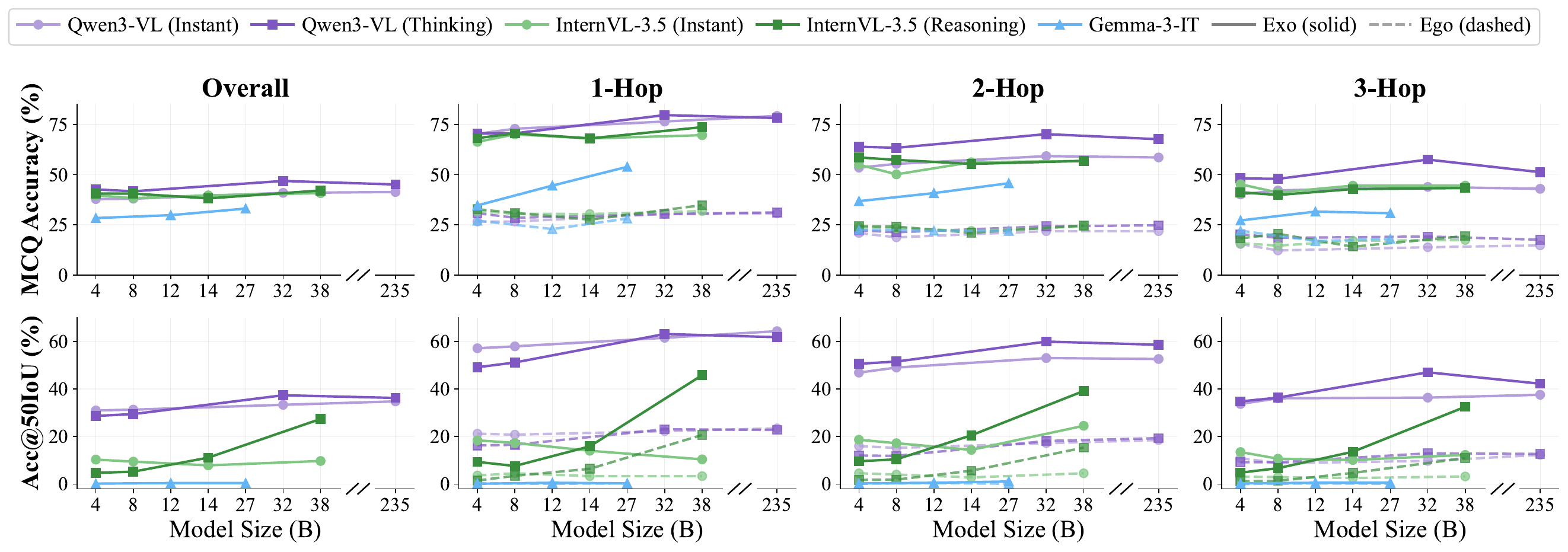}
    \vspace{-0.6cm}
    \caption{\textbf{Effect of model scale on MCQ Accuracy (top) and Acc@50IoU (bottom) across overall and per-hop settings.} Solid lines indicate exo-centric and dashed lines indicate ego-centric performance.}
    \vspace{-0.4cm}
    \label{fig:scale}
\end{figure*}
\begin{figure}[t]
    \centering
        \includegraphics[width=\linewidth]{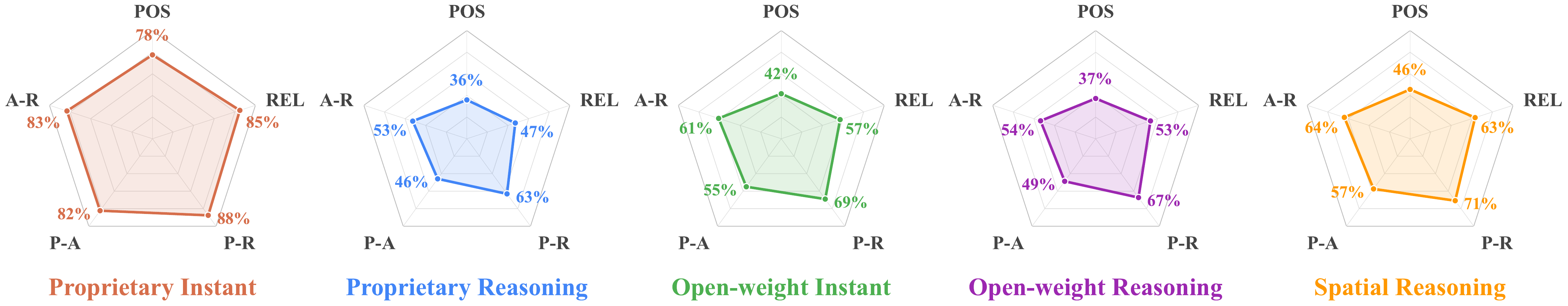}
    \caption{\textbf{Average Error Rates~($\downarrow, \%$) by tag combination across model categories.} A, R, and P denote Attribute, Relation, and Position, respectively.
    }
    \label{fig:error_rate}
\end{figure}

\noindent
\textbf{A7. The Limits of LLM Scaling in Spatial Reasoning.}
Analyzing model scale across three open-weight families (\cref{fig:scale}) reveals that scaling the language backbone alone yields limited gains. While MCQ accuracy shows modest, saturating improvements, Acc@50IoU largely plateaus (\eg, Qwen3-VL) or remains near zero (\eg, Gemma-3-IT). A notable exception is InternVL-3.5 at 38B, which exhibits a sharp grounding jump from 5.2\% to 27.4\%. Crucially, this leap coincides with a significant upgrade in its vision encoder~(from 300M to 6B), whereas models utilizing fixed, small vision encoders (\eg, Qwen3-VL's 400M) saturate early. This stark contrast demonstrates that multi-hop spatial reasoning depends critically on scaling the capacity of visual spatial representations, rather than relying solely on the language component's reasoning capacity.

\noindent
\textbf{A8. Error Analysis on Tag Compositions.}
Analyzing error rates across spatial tag combinations (\cref{fig:error_rate}) reveals that while reasoning models consistently outperform instant models, multi-tag compositions remain a critical bottleneck. Specifically, the \textsc{pos}-\textsc{rel} (\textsc{P-R}) combination incurs significantly higher errors than single-tag settings, highlighting the severe difficulty of jointly handling positional localization and relational comparison. Even specialized spatial reasoning models, which achieve the lowest overall errors, exhibit non-trivial failures on these complex compositions. This underscores that true compositional spatial reasoning remains an unresolved challenge across the current VLM ecosystem.

\begin{table*}[t]
    \centering
    \caption{\textbf{Impact of MultihopSpatial-Train on spatial reasoning across diverse benchmarks.}}
    \label{tab:rl}
    \resizebox{\textwidth}{!}{%
    \begin{tabular}{l|ccc|ccccc}
        \toprule
        & \multicolumn{3}{c|}{\textbf{MultihopSpatial}} & \multicolumn{5}{c}{\textbf{Out-of-Domain}} \\
        \cline{2-9}
        \textbf{Model} & Acc. & Acc@50IoU & avg. IoU & \textbf{BLINK}~\cite{fu2024blink} & \textbf{3DSRBench}~\cite{3dsrbench} & \textbf{OmniSpatial}~\cite{omnispatial25} & \textbf{VSI-Bench}~\cite{vsibench} & \textbf{SpatialMQA}~\cite{spatialmqa}  \\
        \midrule
        Qwen3-VL-4B-Instruct & 37.8 & 31.0 & 69.9 & 82.5 & 56.1 & 42.7 & 62.8 & 39.6 \\
        \textbf{w/ MultihopSpatial-Train} & \textbf{62.9} & \textbf{53.8} & \textbf{72.6} & \textbf{85.3} & \textbf{56.3} & \textbf{43.9} & \textbf{63.2} & \textbf{41.1} \\
        \bottomrule
    \end{tabular}
    }
\end{table*}

\begin{table}[t]
\centering
\caption{\textbf{Vision-Language-Action evaluation on Calvin ABC-D~\cite{mees2022calvin} and Libero~\cite{liu2023libero} benchmark.} Each task~(Task-\#) measures success rate~(\%) on Calvin and Calvin denotes the average number of successfully completed tasks per sequence.}
\scalebox{0.8}{
    \begin{tabular}{lcccccc|c}
    \toprule
    \textbf{VLM backbone} & \textbf{Task-1} & \textbf{Task-2} & \textbf{Task-3} & \textbf{Task-4} & \textbf{Task-5} & \textbf{Calvin} & \textbf{Libero} \\
    \midrule
    Qwen3-VL-4B-Instruct & 92.4 & 81.8 & 74.1 & 66.8 & 59.9 & 3.75 & 35.8 \\
    \textbf{w/ MultihopSpatial-Train} & \textbf{93.0} & \textbf{85.4} & \textbf{79.3} & \textbf{73.2} & \textbf{66.9} & \textbf{3.98} & \textbf{40.0} \\
    \bottomrule
    \end{tabular}
}
\label{tab:vla}
\end{table}

\subsection{MultihopSpatial-Train: Utility as a Training Corpus}
To validate MultihopSpatial as a training resource for improving spatial reasoning in VLMs, we compare our Qwen3-VL-4B-Instruct trained on our Multihop-Train with its baseline on spatial reasoning benchmarks and VLA benchmarks.

\noindent
\textbf{Spatial Reasoning Benchmarks.}
We evaluate the GRPO-trained model on both the in-domain MultihopSpatial benchmark and five out-of-domain spatial reasoning benchmarks in~\cref{tab:rl}.
On the in-domain MultihopSpatial benchmark, GRPO training yields substantial improvements across all three metrics.
The large gains in MCQ Accuracy and Acc@50IoU indicate that the model learns to perform multi-hop spatial reasoning more reliably, while the modest but consistent improvement in Avg.\ IoU suggests that grounding quality also benefits.
More notably, the trained model achieves consistent improvements on all five out-of-domain benchmarks, despite being trained exclusively on MultihopSpatial data.
This suggests that the spatial reasoning capabilities acquired through our dataset generalize beyond the in-domain distribution, transferring to diverse spatial understanding tasks. Additional experimental results with different model scales are described in~\cref{app:model_scale}.

\noindent
\textbf{VLM as backbone for Vision-Language-Action.}
To investigate whether improved spatial reasoning transfers to embodied decision-making, we evaluate the models as VLM backbones using VLM4VLA~\cite{zhang2026vlmvla} on the CALVIN ABC-D~\cite{mees2022calvin} and Libero~\cite{liu2023libero} benchmarks. As shown in~\cref{tab:vla}, the MultihopSpatial-trained model consistently outperforms the baseline. On CALVIN, the average task completion score rises from 3.75 to 3.98, with the performance gap widening as the task chain lengthens (from +0.6\% on Task-1 to +7.0\% on Task-5). This highlights compounding benefits in long-horizon sequential manipulation. Furthermore, the model achieves a +4.2\% absolute improvement (35.8\% $\rightarrow$ 40.0\%) on the Libero benchmark. 
These results demonstrate that spatial reasoning capabilities acquired from MultihopSpatial-Train not only generalize across vision-language benchmarks but also translate into robust robotic policy execution, reinforcing spatial reasoning as a foundational capability for VLAs.

\definecolor{posit}{HTML}{4EA72E}  
\begin{figure}[t]
    \centering
        \includegraphics[width=\linewidth]{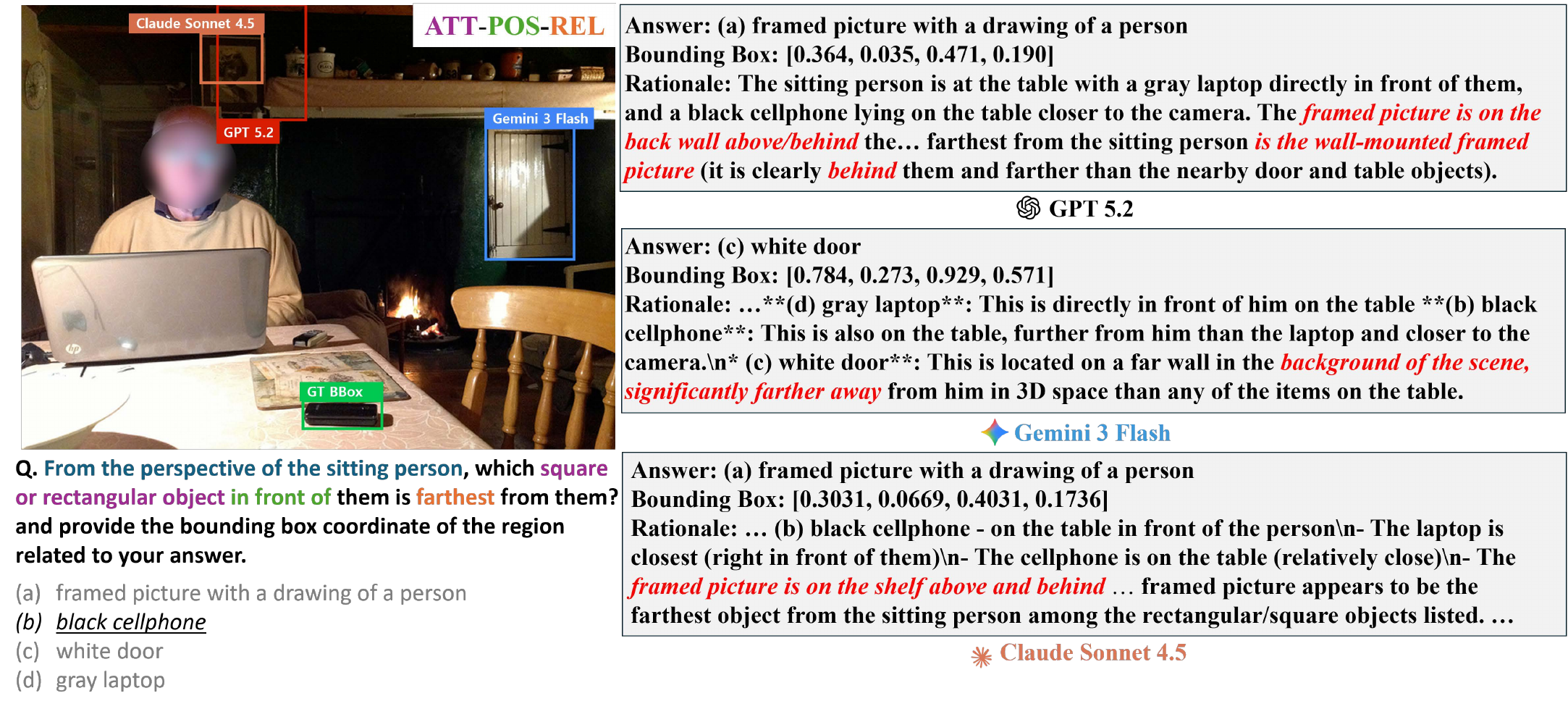}
    \vspace{-0.7cm}
    \caption{
    \textbf{Qualitative failure analysis} on a 3-hop ego-centric question. 
    All three models recognized the position condition \textit{``in front of''}, but failed to incorporate it into their final predictions. Full reasoning traces are provided in Appendix Table~\ref{tab:sup_trace_rect}.
    }
    \vspace{-0.5cm}
    \label{fig:failure_case}
\end{figure}


\subsection{Qualitative Results}

\cref{fig:failure_case} illustrates a representative failure case for a 3-hop ego-centric question, requiring three sequential inferences. The query asks to identify the farthest square or rectangular object  \emph{in front of} the sitting person. However, all three models neglect \textsc{pos}, instead selecting the farthest rectangular object based solely on \textsc{att} and \textsc{rel}. Examining each model's rationale reveals that they explicitly acknowledge the \emph{in front of} constraint during reasoning, yet consistently disregard it in the final answer---demonstrating a failure to maintain intermediate conditions throughout the chain. This highlights the necessity of our compositional evaluation: an error at any single hop inevitably propagates to an incorrect final prediction. Additional qualitative results are provided in~\cref{app:qual}.
\section{Conclusion}
We introduce \textbf{MultihopSpatial}, a comprehensive dataset for evaluating and enhancing multi-hop, compositional spatial reasoning in VLMs. To ensure evaluations explicitly reflect the visual grounding vital for Vision-Language-Action (VLA) deployment, we propose \textbf{Acc@50IoU}, a complementary grounded metric that simultaneously evaluates correct answer selection and precise spatial localization. While our extensive evaluation of 37 models reveals that complex spatial reasoning remains a formidable challenge, we demonstrate that RL post-training on our dedicated corpus effectively mitigates this issue. Ultimately, MultihopSpatial drives consistent improvements in both intrinsic VLM reasoning and downstream VLA task execution. We hope our dataset and grounded evaluation paradigm will catalyze future research in Embodied AI, particularly in scaling up spatial priors for real-time, dynamic robotic control in the wild.

\section{Acknowledgments}
This work was supported by Institute of Information \& communications Technology Planning \& Evaluation (IITP) grant funded by the Korea government (MSIT) (No. 2022-0-00871, Development of AI Autonomy and Knowledge Enhancement for AI Agent Collaboration, 90\%) and (No.2019-0-00075, Artificial Intelligence Graduate School Program(KAIST), 10\%).

%
%
\bibliographystyle{splncs04}
\bibliography{main}

\clearpage
\appendix                 
\renewcommand{\theHsection}{A.\thesection}

\crefalias{section}{appendix} 
\crefname{appendix}{App.}{Apps.} 
\Crefname{appendix}{Appendix}{Appendices} 

\providecommand{\authcount}[1]{}

\vspace*{2em}
\begin{center}
    {\Huge \bfseries Appendices \par}
\end{center}
\vspace*{2em}

\startcontents[sections]
\printcontents[sections]{}{1}{\setcounter{tocdepth}{2}}

\clearpage

\section{Training Details}\label{app:train}

\subsection{Training for VLM}
We use Qwen3-VL-4B-Instruct~\cite{qwen3vl} as the base policy model.
We apply LoRA~\cite{lora} to the LLM backbone while keeping the base LLM weights, vision encoder, and vision-language merger frozen.
The learning rate for the LoRA adapters is set to $5 \times 10^{-5}$.
For GRPO, we sample $G = 4$ candidate outputs per question with a maximum completion length of $2{,}048$ tokens and a maximum prompt length of $1{,}024$ tokens.
Training is conducted for $10$ epochs using the AdamW optimizer~\cite{adamw} ($\beta_1 = 0.9$, $\beta_2 = 0.999$) with a cosine learning rate schedule, $3\%$ warmup ratio, weight decay of $0.1$, and learning rate of  $5 \times 10^{-5}$.
The effective batch size is $128$ ($8$ GPUs $\times$ $16$ prompts per GPU).
We use DeepSpeed ZeRO Stage-2 with BF16 mixed precision and gradient checkpointing.
Our implementation builds on the TRL library~\cite{trl}.
All reward weights are set to $\alpha = \beta = 1$ in Eq.~\eqref{eq:total_reward}.

\subsection{Training for VLA}
For Vision-Language-Action evaluation, we adopt VLM4VLA~\cite{zhang2026vlmvla} which enables fair comparison of different VLM backbones under the same training pipeline and evaluation protocol. Specifically, we use Qwen3-VL-4B-Instruct as the VLM backbone. We conduct CALVIN~\cite{mees2022calvin} experiments on the ABC→D split, which contains approximately 1.05M training samples from 147 episodes across three environments (A, B, C). We follow the VLM4VLA~\cite{zhang2026vlmvla} settings, which insert learnable action tokens into the VLM's input embedding sequence and append an MLP-based action head (FCDecoder) to decode continuous 7-DoF actions (6-DoF end-effector pose + 1-DoF gripper). We train with the Adam optimizer using a learning rate of $2\times10^{-5}$ with cosine annealing (minimum LR scale of 0.01) and a linear warmup over the first 0.25 epochs. We employ BFloat16 mixed precision and DeepSpeed ZeRO Stage 2 for distributed training across 8 NVIDIA A100 (80GB) GPUs with gradient clipping of 1.0. The entire model—including the vision encoder, language model backbone, and action head—is fine-tuned end-to-end without LoRA. Action labels are normalized to the range $[-0.65, 0.65]$. The action head employs a Huber loss for the 6-DoF arm actions and binary cross-entropy for the gripper action, with a gripper loss weighting ratio of 0.01. We use a per-GPU batch size of 16 (effective batch size of 128) and train for 2 epochs (16,466 optimizer steps).

For LIBERO~\cite{liu2023libero} experiments, we evaluate on the LIBERO-10 suite, which comprises 10 long-horizon tabletop manipulation tasks with 50 demonstrations per task. We use the same VLM4VLA framework with Qwen3-VL-4B-Instruct as the VLM backbone and the FCDecoder action head. Training employs the Adam optimizer with a learning rate of $2\times10^{-5}$, cosine annealing (minimum LR scale of 0.01), and linear warmup over the first 0.25 epochs. We use BFloat16 mixed precision and DeepSpeed ZeRO Stage 2 across 8 NVIDIA A100 (80GB) GPUs with gradient clipping of 1.0, and fine-tune the entire model end-to-end without LoRA. Action labels are normalized to $[-0.65, 0.65]$ with the same loss configuration (Huber loss for arm actions, binary cross-entropy for the gripper, gripper loss ratio of 0.01). Unlike the CALVIN setup, we use an action chunk size of 4 (versus 10 for CALVIN) and apply gradient accumulation with a factor of 4, resulting in a per-GPU batch size of 16 with an effective batch size of 512. We train for 15K optimizer steps and select the best checkpoint based on evaluation performance.

\begin{table}[t]
\centering
\caption{\textbf{Ablation Study: RL vs. SFT Post-training on MultihopSpatial-Train.}}
\begin{tabular}{lccc}
\toprule
\textbf{Model} & \textbf{MCQ Acc.} & \textbf{Acc@50IoU} & \textbf{avg. IoU} \\
\midrule
Qwen3-VL-4B-Instruct & 28.9 & 23.3 & 67.4 \\
$+$ \textbf{SFT} & 58.1 & 49.0 & 71.5 \\
$+$ \textbf{GRPO ($R_{\text{mcq}}$)} & \textbf{61.8} & 51.6 & 71.4 \\
$+$ \textbf{GRPO ($R_{\text{mcq}} + R_{\text{bbox}}$)} & 61.1 & \textbf{52.3} & \textbf{72.4} \\
\bottomrule
\end{tabular}
\label{tab:ablation}
\end{table}
\section{Additional Experiments}

\noindent\textbf{Ablation Setup.} 
Unless otherwise specified, all ablation studies in this section follow the same training configuration as our main experiments detailed in~\cref{app:train}, with one key modification: the models are trained for 5 epochs instead of 10. This reduced training duration allows for efficient exploration of various hyperparameters. To ensure a strictly fair comparison in the subsequent experiments, both the SFT and GRPO (RL) models are trained for 5 epochs using an identical learning rate of $5 \times 10^{-5}$.

\subsection{Ablation Study: RL vs. SFT Post-training}

\noindent
\textbf{SFT with bounding box prediction.}
To enable the SFT model to jointly predict the multiple-choice answer and the bounding box, we adopt a response augmentation strategy within the standard next-token prediction (NTP) framework. Specifically, during data loading, we transform each ground-truth response from the original MCQ-only format (\eg, ``\texttt{(b) green round container}'') into a structured format: ``\texttt{Answer: (b) green round container\textbackslash nBounding Box: [876, 506, 940, 604]}''. The ground-truth bounding box, originally in pixel coordinates $[x, y, w, h]$, is converted to the $[x_1, y_1, x_2, y_2]$ format on a 0--1000 normalized scale following Qwen3-VL~\cite{qwen3vl}. This allows the model to learn spatial localization purely through the language modeling objective, without architectural modifications or auxiliary detection heads.

\noindent
\textbf{Results and analysis.}
As shown in~\cref{tab:ablation}, SFT provides a substantial improvement over the base model across all metrics (+29.2\% MCQ Acc., +25.7\% Acc@50IoU), confirming that task-specific fine-tuning is essential for multi-hop spatial reasoning. Building upon this, GRPO with only the MCQ reward ($R_{\text{mcq}}$) further improves MCQ accuracy to 61.8\% (+3.7\%) and Acc@50IoU to 51.6\% (+2.6\%), demonstrating that RL can discover reasoning strategies beyond standard supervised imitation.

The most notable finding emerges when comparing SFT and GRPO ($R_{\text{mcq}} + R_{\text{bbox}}$) for visual grounding. Although SFT learns to generate coordinates through NTP, this implicit supervision treats coordinate tokens identically to natural language tokens, providing no geometric feedback on localization quality. In contrast, GRPO with an explicit bounding box reward ($R_{\text{bbox}}$) directly optimizes for IoU-based accuracy, yielding the highest Acc@50IoU (52.3\%) and average IoU (72.4\%). This suggests that explicit reward-based optimization is vastly more effective than implicit next-token prediction for spatial grounding, as the reward signal provides a direct measure of geometric correctness that token-level cross-entropy loss cannot capture.

Finally, comparing the two GRPO variants reveals a natural trade-off: adding $R_{\text{bbox}}$ causes a marginal decrease in MCQ accuracy (61.8\% $\rightarrow$ 61.1\%) while improving localization metrics (51.6\% $\rightarrow$ 52.3\% Acc@50IoU). When the reward function includes a bounding box component, policy optimization allocates part of its capacity toward improving localization, slightly shifting the focus away from pure answer selection. Nevertheless, the MCQ accuracy drop is minimal ($-0.7\%$), while the localization gains are more pronounced, indicating that the combined reward achieves an optimal balance between reasoning and grounding.

\begin{table}[t]
  \centering
  \caption{\textbf{Ablation Study: $\alpha$ \& $\beta$ in Reward Function.}}
    \begin{tabular}{lccc}
    \toprule
    $\alpha \ \& \ \beta$ & MCQ Acc. & Acc@50IoU & avg. IoU \\
    \midrule
    1 \& 1 & \textbf{61.1} & \textbf{52.3} & 72.4 \\
    1 \& 2 & 60.6 & 51.1 & 71.6 \\
    2 \& 1 & 52.5 & 45.3 & \textbf{72.8} \\
    2 \& 2 & 59.9 & 51.8 & 72.6 \\
    \bottomrule
    \end{tabular}
    \label{tab:ablation_alpha_beta}
\end{table}
\subsection{Ablation Study on $\alpha$ and $\beta$ in the Reward Function}
We conduct an ablation study to explore the impact of the hyperparameters $\alpha$ and $\beta$ in the reward function~(\cref{eq:total_reward}). As shown in~\cref{tab:ablation_alpha_beta}, the configuration of $\alpha=1$ and $\beta=1$ achieves the optimal balance, yielding the highest MCQ Accuracy (61.1\%) and Acc@50IoU (52.3\%). Interestingly, increasing $\alpha$ disproportionately ($\alpha=2, \beta=1$) maximizes the average IoU (72.8) but causes a severe degradation in MCQ Accuracy (52.5\%). This reveals a clear trade-off: over-weighting one reward component forces the policy to over-optimize for spatial localization at the severe expense of logical reasoning. Furthermore, scaling both weights equally ($\alpha=2, \beta=2$) largely restores performance compared to the imbalanced settings, demonstrating that the relative ratio between the rewards is more critical than their absolute magnitudes. Based on these findings, we adopt $\alpha=1$ and $\beta=1$ as our default configuration.

\begin{table}[t]
\centering
\caption{\textbf{Training corpus comparison with different model scales.} ID: In-domain. OOD Avg. denotes the mean performance across the five OOD benchmarks.}
\setlength{\tabcolsep}{2.8pt}
\renewcommand{\arraystretch}{0.9}
\resizebox{\columnwidth}{!}{
\begin{tabular}{l | cc | cccccc}
\toprule
\multirow{2}{*}{Model} & \multicolumn{2}{c|}{MultihopSpatial (ID)} & \multicolumn{6}{c}{Out-Of-Domain (OOD)} \\
\cmidrule{2-9}
 & Acc. & Acc@50 & BLINK & 3DSR & Omni & VSI & SpaMQA & OOD Avg. \\
\midrule
Qwen3-VL-4B & 37.8 & 31.0 & 82.5 & 56.1 & 42.7 & 62.8 & 39.6 & 56.7 \\
\hspace{1em}+ OmniSpatial & 37.1 & 30.6 & 78.3 & \textbf{56.8} & 43.1 & \textbf{63.7} & \textbf{42.8} & 56.9 \\
\hspace{1em}+ \textbf{MultihopSpatial} & \textbf{62.9} & \textbf{53.8} & \textbf{85.3} & 56.3 & \textbf{43.9} & 63.2 & 41.1 & \textbf{58.0} \\
\midrule
Qwen3-VL-8B & 38.0 & 31.3 & \textbf{85.3} & 56.3 & 47.5 & 62.5 & 39.1 & 58.2 \\
\hspace{1em}+ OmniSpatial~\cite{omnispatial25} & 38.1 & 31.0 & 79.7 & 56.4 & 47.1 & \textbf{63.0} & \textbf{43.7} & 58.0 \\
\hspace{1em}+ \textbf{MultihopSpatial} & \textbf{61.0} & \textbf{51.5} & 83.9 & \textbf{59.5} & \textbf{48.7} & 62.6 & 41.3 & \textbf{59.2} \\
\midrule
Qwen3-VL-32B & 41.0 & 33.4 & 89.5 & 61.2 & 49.6 & 66.9 & 43.5 & 62.1 \\
\hspace{1em}+ OmniSpatial~\cite{omnispatial25} & 41.2 & 33.7 & \textbf{90.2} & 61.1 & 50.5 & \textbf{67.3} & 47.6 & 63.3 \\
\hspace{1em}+ \textbf{MultihopSpatial} & \textbf{67.2} & \textbf{56.9} & 88.1 & \textbf{62.7} & \textbf{55.1} & 64.5 & \textbf{53.0} & \textbf{64.7} \\
\bottomrule
\end{tabular}}
\label{tab:rebuttal}
\end{table}
\subsection{Comparison with other train corpora:OmniSpatial~\cite{omnispatial25}}\label{app:tab_omni_comparison}
We compare against OmniSpatial-Train ($6,885$ vs.\ ours $6,791$) under the same training setup (\cref{tab:rebuttal}). MultihopSpatial-Train achieves a higher OOD Avg. across all models, confirming that our improvements stem from superior data quality rather than quantity. 

\subsection{Generalization to model scale.}\label{app:model_scale}
Although modest in size, our 6.8k corpus is fully human-annotated and rigorously verified, prioritizing highly reliable, high-density reasoning signals over sheer volume. To address scalability, RL post-training on Qwen3-VL-8B and 32B consistently improves both ID and OOD performance over base models~(\cref{tab:rebuttal}), confirming our approach effectively scales to larger architectures and generalizable supervision.


\begin{figure}[h]
  \centering
  \includegraphics[width=0.8\linewidth]{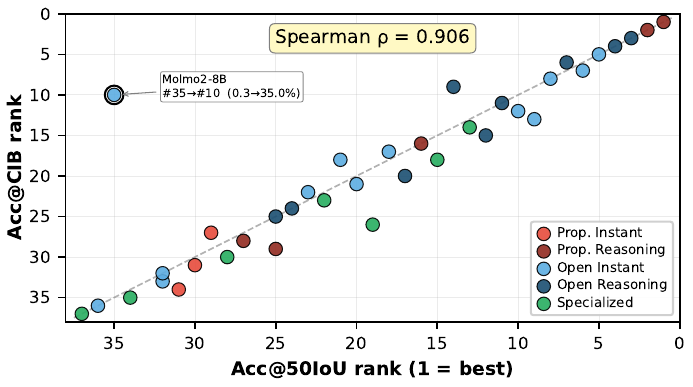}
  \caption{\textbf{Rank Correlations between metrics.}}
  \label{fig:rebuttal-corr}
\end{figure}


\subsection{Flexibility of the evaluation metrics.}
We re-evaluate all 37 models with two complementary metrics: a looser threshold \textbf{Acc@25IoU} and \textbf{Acc@CIB}~(\textit{Center-In-Box}), a pointing-game metric requiring the MCQ to be correct and the predicted box's center to lie in the GT box. Both correlate tightly with Acc@50IoU (Spearman $\rho{=}0.986$ and $0.906$; \cref{fig:rebuttal-corr} shows the latter), with only $2$ and $3$ of $37$ models showing $|\Delta\text{rank}|{\geq}5$---confirming that Acc@50IoU rarely penalizes models unfairly. Notably, Molmo2-8B (\#35$\rightarrow$\#10) localizes the correct region with imprecise box geometry. We expect Acc@CIB to serve as a \textit{complementary} metric, capturing coarse-but-correct localization that strict IoU may overlook.

\section{Limitation}
While MultihopSpatial comprehensively evaluates and enhances multi-hop spatial reasoning, a few limitations remain. First, our RL post-training experiments were primarily conducted on a compact model (\eg, Qwen3-VL-4B-Instruct~\cite{qwen3vl}). Exploring how the compounding benefits of this spatial post-training scale with much larger VLMs is an important direction we leave for future work. Second, our current benchmark and training corpus are constructed exclusively from static images. Extending our compositional reasoning framework to spatial-temporal environments (\eg, video-based multi-hop compositional reasoning) represents a critical next step. Third, MultihopSpatial probes foundational compositional reasoning, not universal embodied scenarios. COCO and PACO-Ego4D provide heavily cluttered daily scenes with complex, entangled spatial relations. While specialized domains (industrial, warehouse) demand identical reasoning logic, the lack of their specific visual features is a limitation. We believe addressing these areas in future research will further advance the foundational spatial intelligence of embodied agents.

\section{System Prompt}\label{app:prompt}

\begin{tcblisting}{
  title=Instruction for evaluation on MultihopSpatial,
  colback=black!5!white,
  colframe=black!50!black,
  listing only, % 아래에 소스코드를 중복 출력하지 않고 박스 안에 내용만 보여줌
  fontupper=\ttfamily % 코드 느낌이 나도록 고정폭 글꼴 적용
}
{Question}

Please respond in the following format:
Answer: (your choice, e.g., "(a) object name")
Bounding Box: {{"bbox_2d": [x1, y1, x2, y2]}}

Important: Use NORMALIZED coordinates (0.0 to 1.0).
Example: {{"bbox_2d": [0.25, 0.1, 0.75, 0.8]}}
\end{tcblisting}

\section{Annotation and Verification Guidelines}\label{app:anno}

\subsection{Annotation}
\begin{figure}[t]
    \centering
        \includegraphics[width=\linewidth]{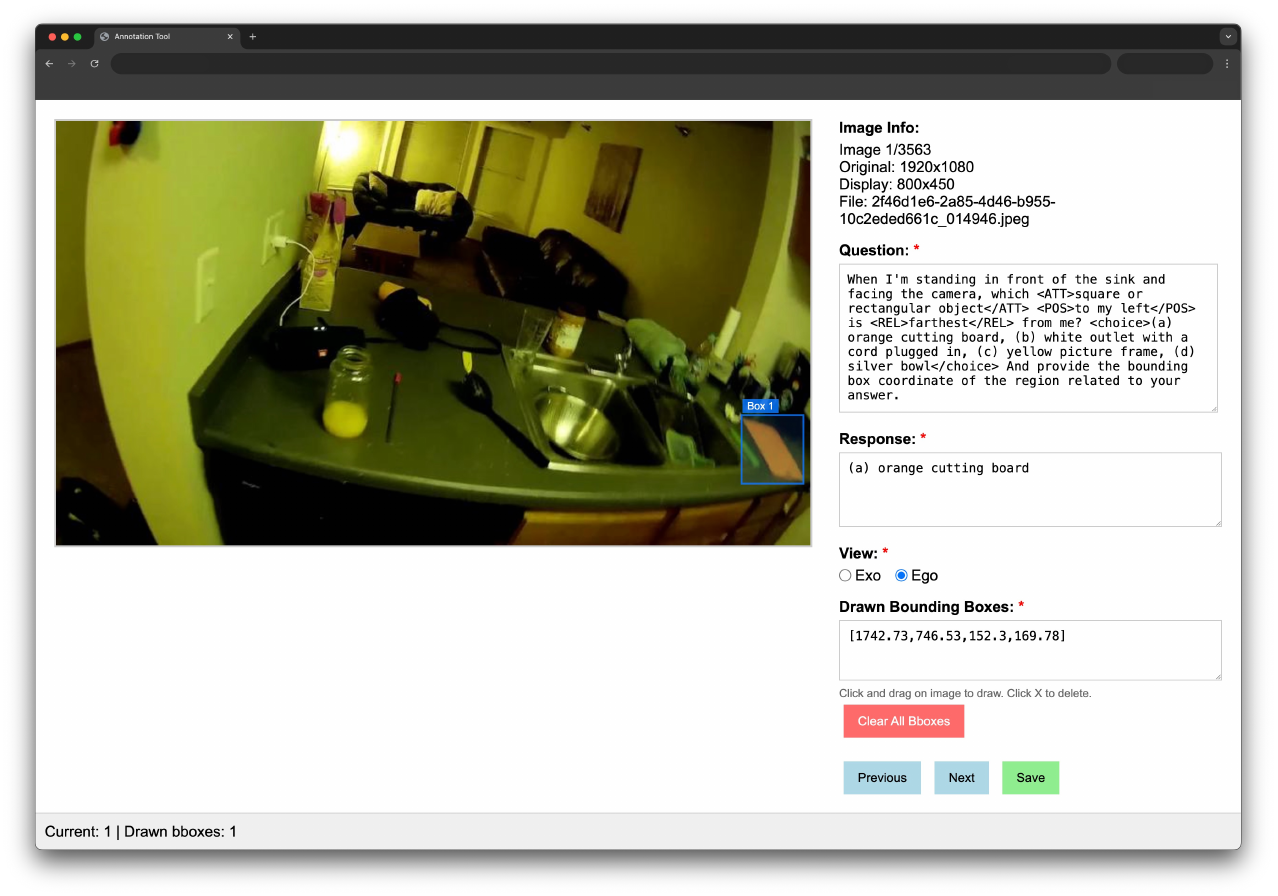}
\caption{\textbf{Screenshot of the data annotation interface.} Annotators write a question-answer pair, select a view type, and draw a bounding box.}
    \label{fig:supp_annotation_tool}
\end{figure}

The QA pairs in MultihopSpatial were manually designed by ten trained annotators based on 3,563 images curated from COCO~\cite{COCO} and PACO-Ego4D~\cite{paco}. Each QA pair follows an N-hop structure composed of three spatial reasoning categories --- Attribute (\textsc{att}), Position (\textsc{pos}), and Relation (\textsc{rel}) --- spanning 1-hop to 3-hop complexities across exo-centric and ego-centric viewpoints. Each pair consists of a question, four multiple-choice options, and a response, along with a bounding box annotation for the target object. \cref{fig:supp_annotation_tool} shows the annotation interface.

\noindent\textbf{Viewpoint Definition.}
\textbf{Exo-centric} view refers to a third-person perspective from an outside observer surveying the entire scene. \textbf{Ego-centric} view refers to the first-person perspective of either a real person appearing in the image or a virtual person placed within the image space. For the ego-centric view, the frontal direction of the perspective subject must be identifiable. Ego-centric view is not permitted when the image is taken from an overhead angle or when the subject lacks a notion of front and back (e.g., a ball or a plate). Inanimate objects cannot serve as the perspective subject.

\noindent\textbf{N-Hop Spatial Reasoning Structure.}
Each QA pair is structured as a compositional problem combining \textsc{att}, \textsc{pos}, and \textsc{rel}, as summarized in~\cref{tab:supp_tab_spatial_categories}. The hop count equals the number of categories in the question.
\begin{table}[h]
\caption{Definitions and examples of the three spatial reasoning categories used in MultihopSpatial.}
\label{tab:supp_tab_spatial_categories}
\resizebox{\textwidth}{!}{
\begin{tabular}{lll}
\toprule
\textbf{Category} & \textbf{Definition} & \textbf{Examples} \\
\midrule
ATT & Visual properties of an object (color, shape, state, etc.) & red object, sitting person, round object \\
POS & Spatial location of an object & left side of the image, in front of the man, on the desk \\
REL & Spatial relationship between objects (comparative, superlative) & highest, closest, farthest \\
\bottomrule
\end{tabular}
}
\end{table}
1-Hop questions use either \textsc{pos} or \textsc{rel} alone; 2-Hop questions combine two categories (\textsc{att+pos}, \textsc{att+rel}, or \textsc{pos+rel}); and 3-Hop questions incorporate all three categories (\textsc{att+pos+rel}). Distance (\textsc{rel}) is judged based on 3D spatial depth rather than 2D pixel distance in the image.

\noindent\textbf{QA Construction Guidelines.}
\textbf{Question} must include the reasoning categories corresponding to the designated hop count. Ego-centric view questions must begin with a subject-establishing phrase: \textit{"From the perspective of [person description],"} for a real person, or \textit{"When I'm standing [location/situation] and facing the camera,"} for a virtual person.
\textbf{Choice} consists of four multiple-choice items, each referring to a single object. Objects must be present in the image and distinguishable by their visual characteristics. If the \textsc{att} condition is color, color descriptors must be excluded from the options. If the question targets a specific category (e.g., person, vehicle), all options must belong to that same category. The options must not be eliminable from the question alone.
\textbf{Response} is the option that is visually accurate given the question.

\subsection{Verification}
\begin{figure}[t]
    \centering
        \includegraphics[width=\linewidth]{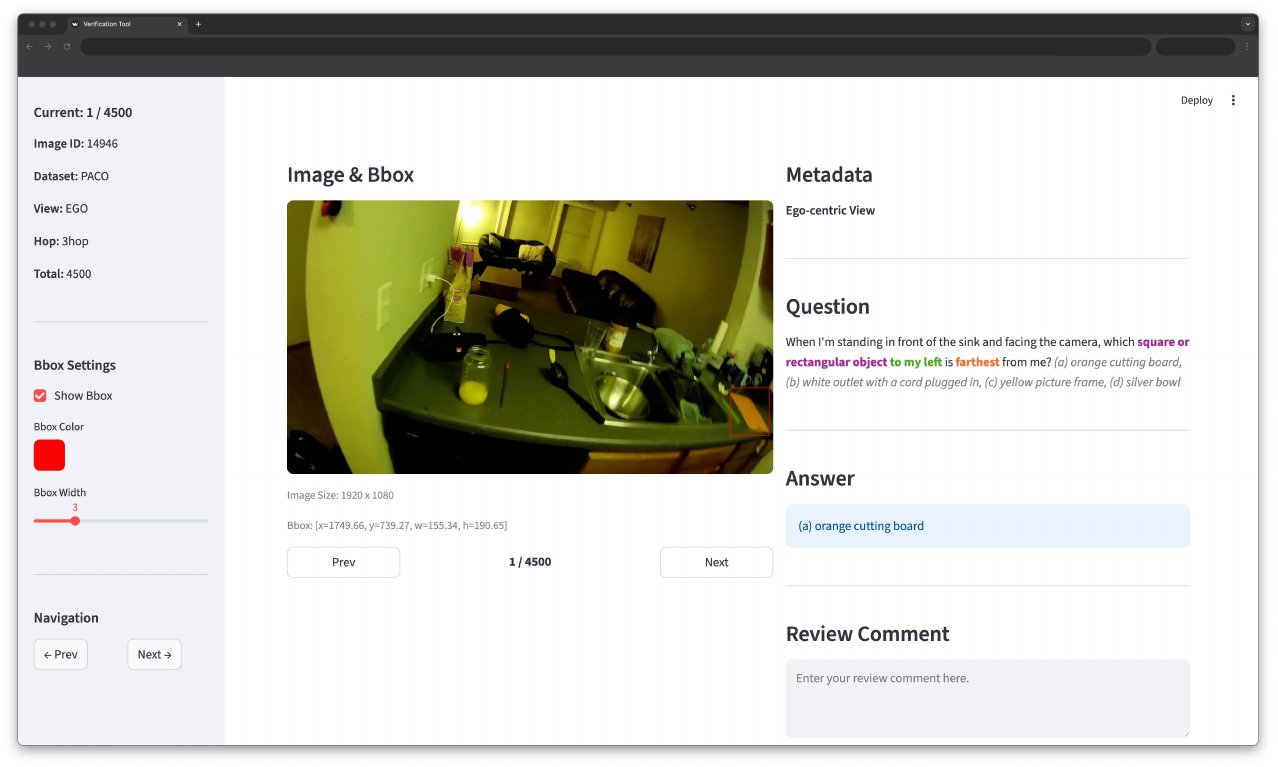}
\caption{\textbf{Screenshot of the data verification interface.} Reviewers verify samples based on the criteria and provide corrections and comments.}
    \label{fig:supp_verification_tool}
\end{figure}

Each QA pair underwent a rigorous multi-stage verification process consisting of three independent rounds. In the first round, annotators checked whether (i) all objects in the question and options exist in the image, (ii) the bounding box corresponds to the correct answer, (iii) the question follows the designated template, and (iv) the answer is visually accurate. In the second round, a different group of annotators cross-examined the samples against the first-round criteria and conducted additional checks for grammatical naturalness and whether any option could be eliminated without viewing the image. A QA pair was rejected if it failed at either of these two initial stages, including cases where the question omitted a required reasoning category, any option referred to more than one object, or the bounding box was missing or incorrect. 
Finally, in the third round, three independent verifiers conducted a conclusive review to ensure unanimous agreement. They strictly verified that all option entities exist in the image, the bounding-box annotation precisely matches the target, and the labeled answer is uniquely supported. 
Following this exhaustive three-round protocol, we measured the inter-annotator agreement, achieving a remarkable Krippendorff's $\alpha = 0.90$. Given the inherent complexity and potential subjectivity involved in evaluating multi-hop and compositional spatial reasoning, this high score is significant. It strongly demonstrates that our rigorous multi-stage filtering successfully eliminated the ambiguity, hallucination, and noise typically found in AI-generated data. Consequently, this high consensus guarantees that our finalized benchmark provides a strictly objective, highly reliable, and unambiguous gold standard for evaluating the true spatial reasoning capabilities of vision-language models.
\cref{fig:supp_verification_tool} shows the interfaces used throughout the verification process.

\section{Qualitative Results}\label{app:qual}
\begin{figure}
    \centering
        \includegraphics[width=\linewidth]{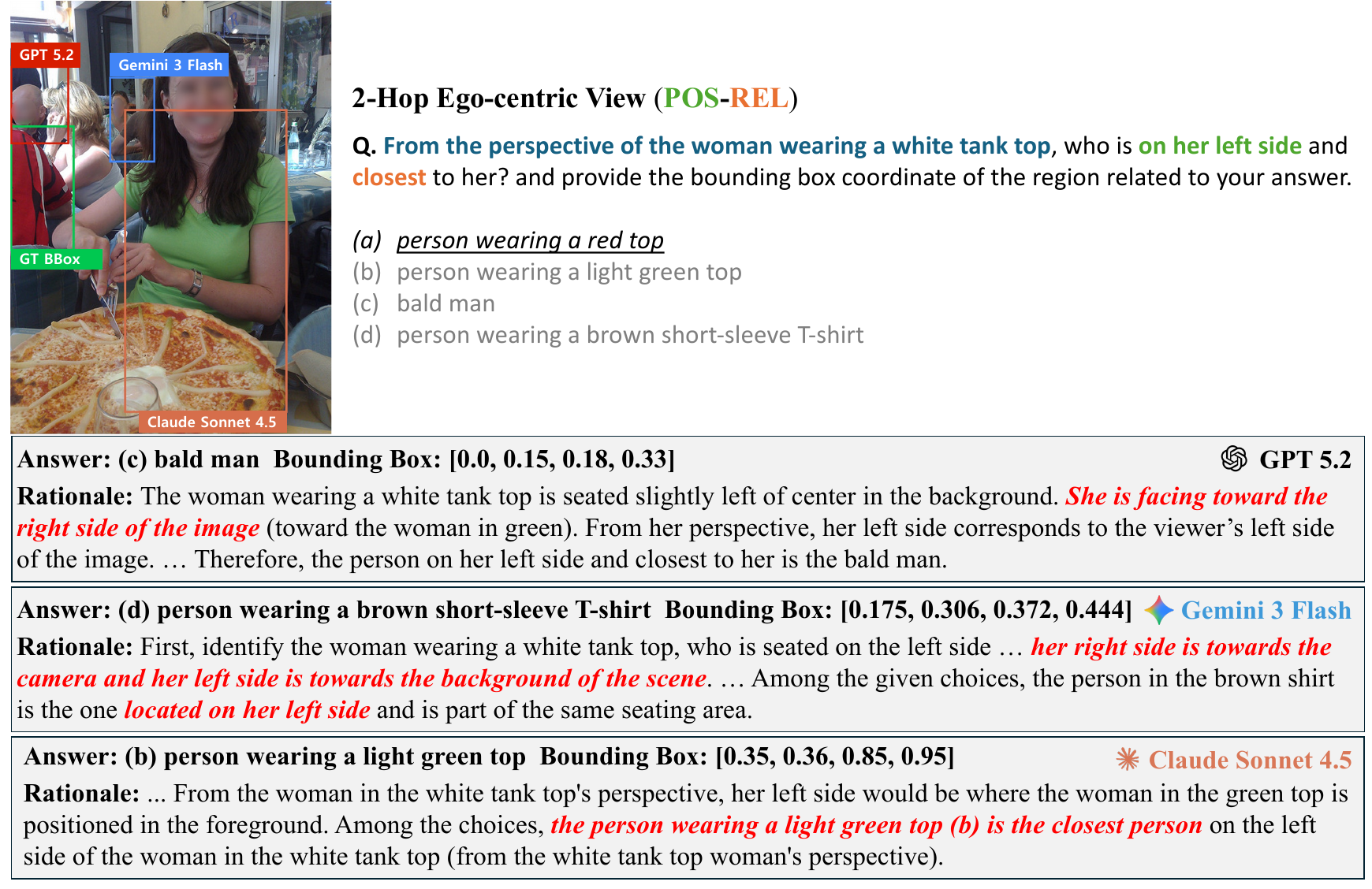}
    \vspace{-0.5cm}
    \caption{\textbf{Qualitative failure analysis on a 2-hop ego-centric question.} GPT, Gemini, and Claude fail on perspective, position, and relation, respectively.}
    \label{fig:qual_ego2hop1}
\end{figure}

\begin{figure}
    \centering
        \includegraphics[width=\linewidth]{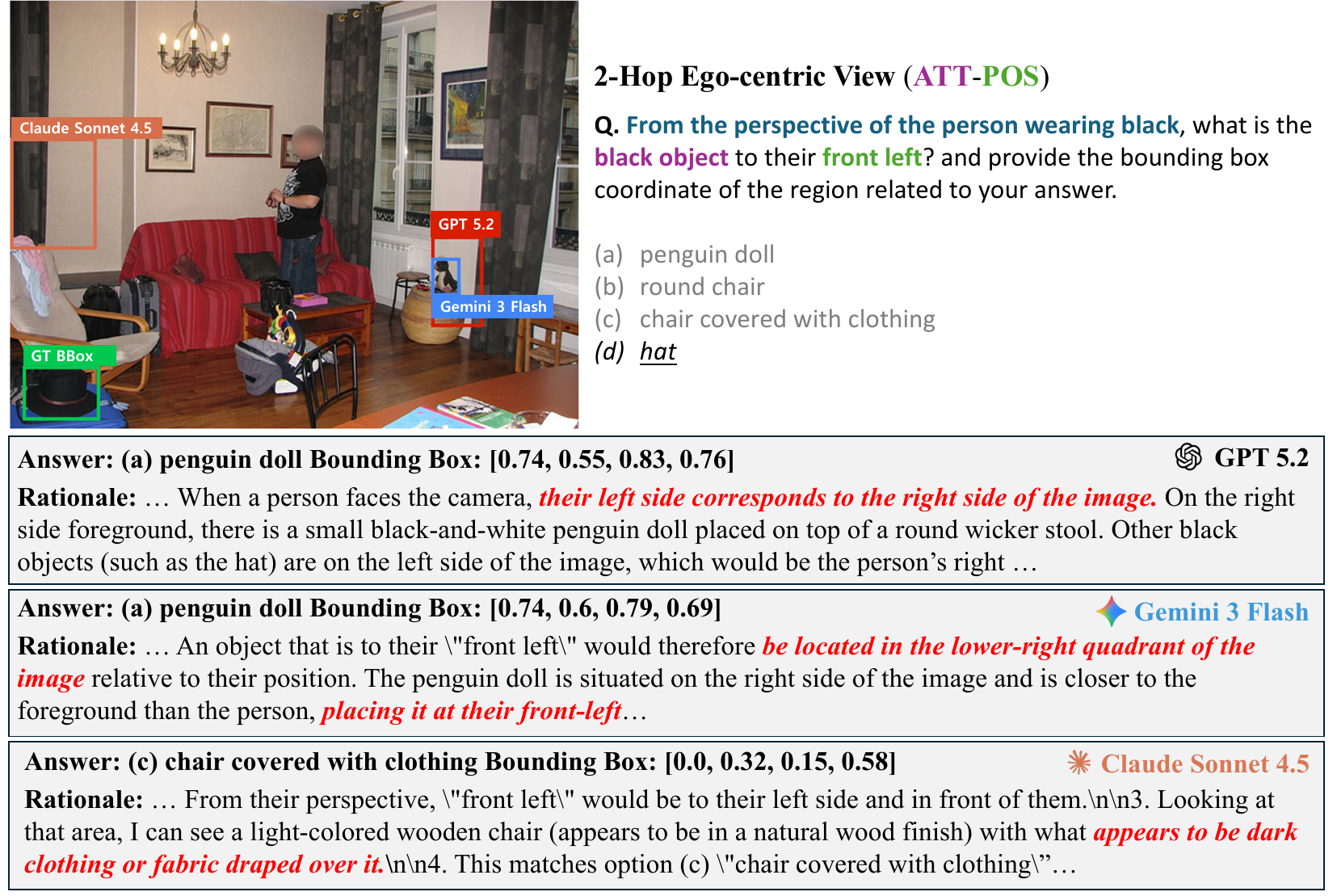}
    \vspace{-0.5cm}
    \caption{\textbf{Qualitative failure analysis on a 2-hop ego-centric question.} GPT and Gemini fail on position recognition, whereas Claude fails on attribute recognition.}
    \label{fig:qual_ego2hop2}
\end{figure}

\begin{figure}
    \centering
        \includegraphics[width=\linewidth]{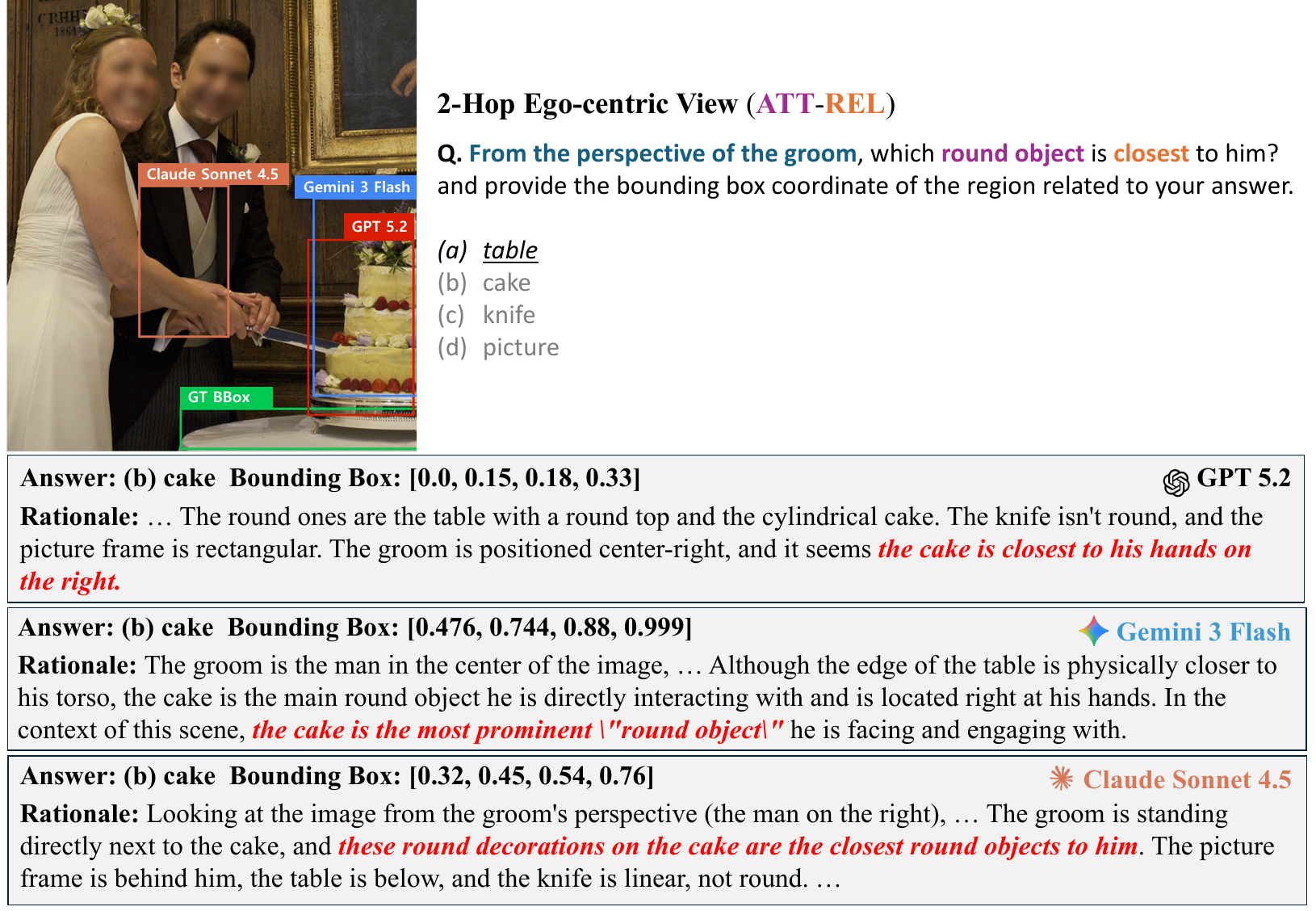}
    \vspace{-0.5cm}
    \caption{\textbf{Qualitative failure analysis on a 2-hop ego-centric question.} All three models fail on relation recognition.}
    \label{fig:qual_ego2hop3}
\end{figure}

\begin{figure}[t]
    \centering
        \includegraphics[width=\linewidth]{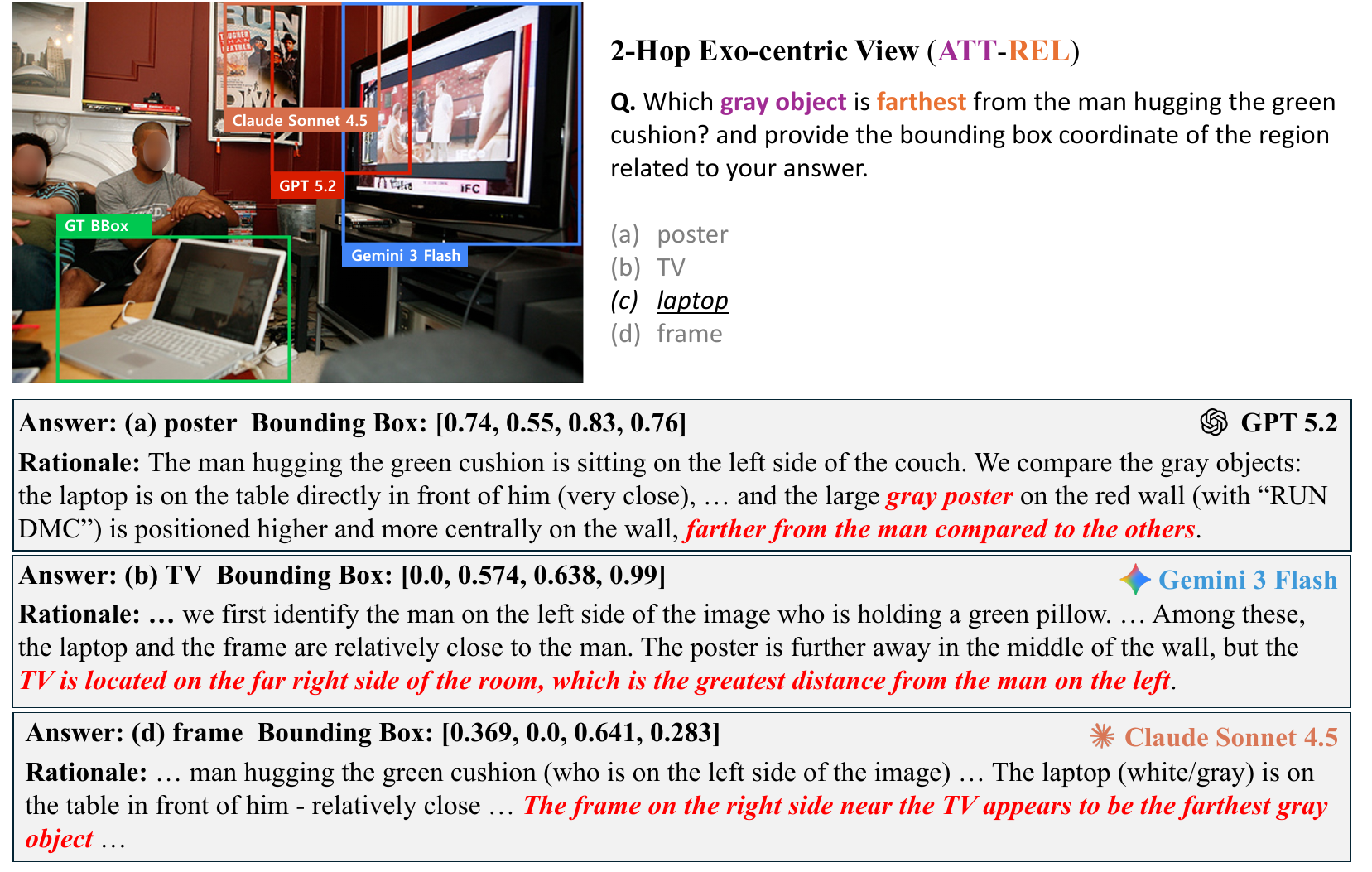}
    \caption{\textbf{Qualitative failure analysis on a 2-hop exo-centric question.} All three models fail to recognize the attribute condition.}
    \label{fig:qual_exo2hop1}
\end{figure}

\begin{figure}[t]
    \centering
        \includegraphics[width=\linewidth]{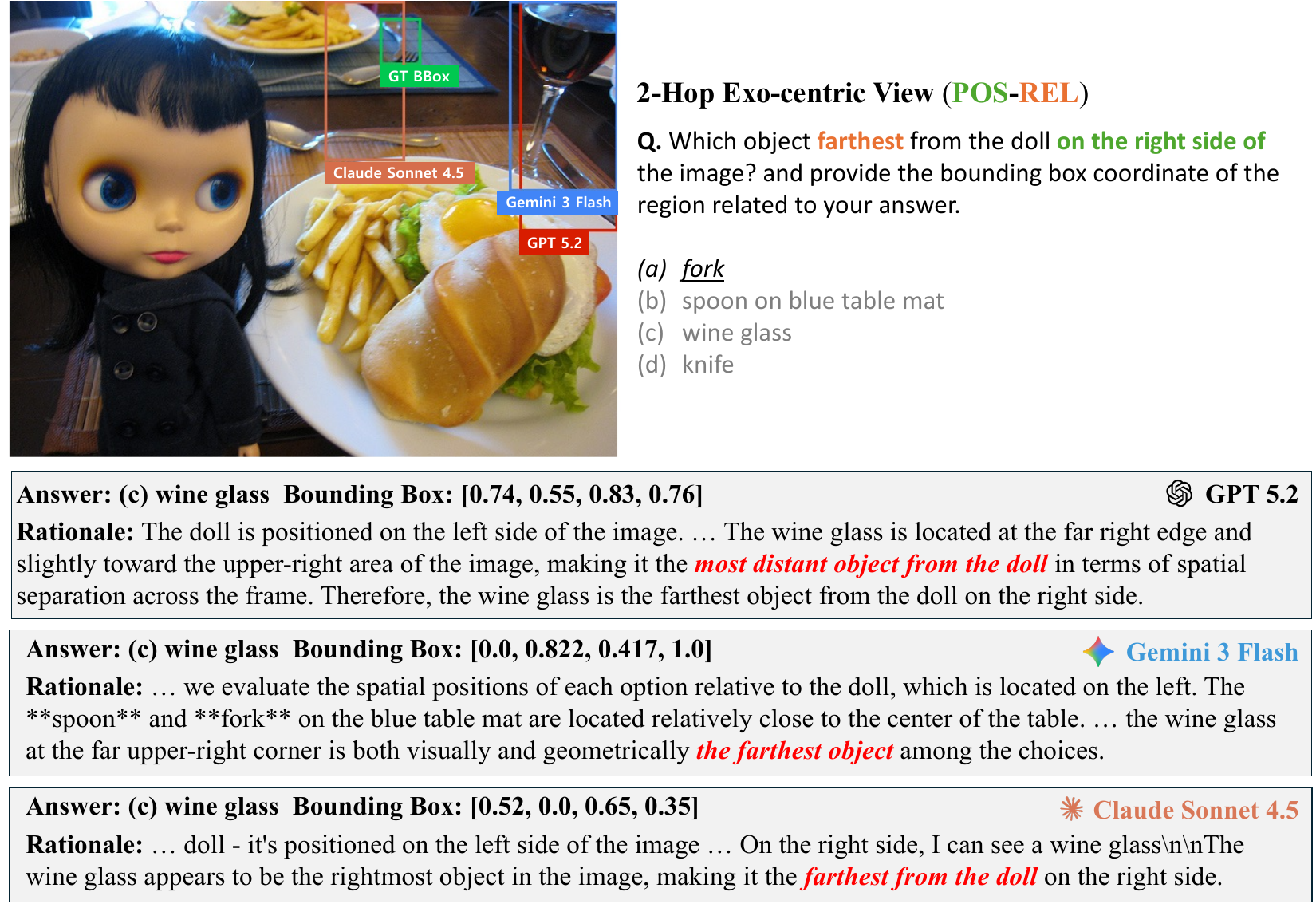}
    \caption{\textbf{Qualitative failure analysis on a 2-hop exo-centric question.} All three models fail to correctly determine the spatial relation based on the 3D distance.}
    \label{fig:qual_exo2hop2}
\end{figure}

\begin{figure}[t]
    \centering
        \includegraphics[width=\linewidth]{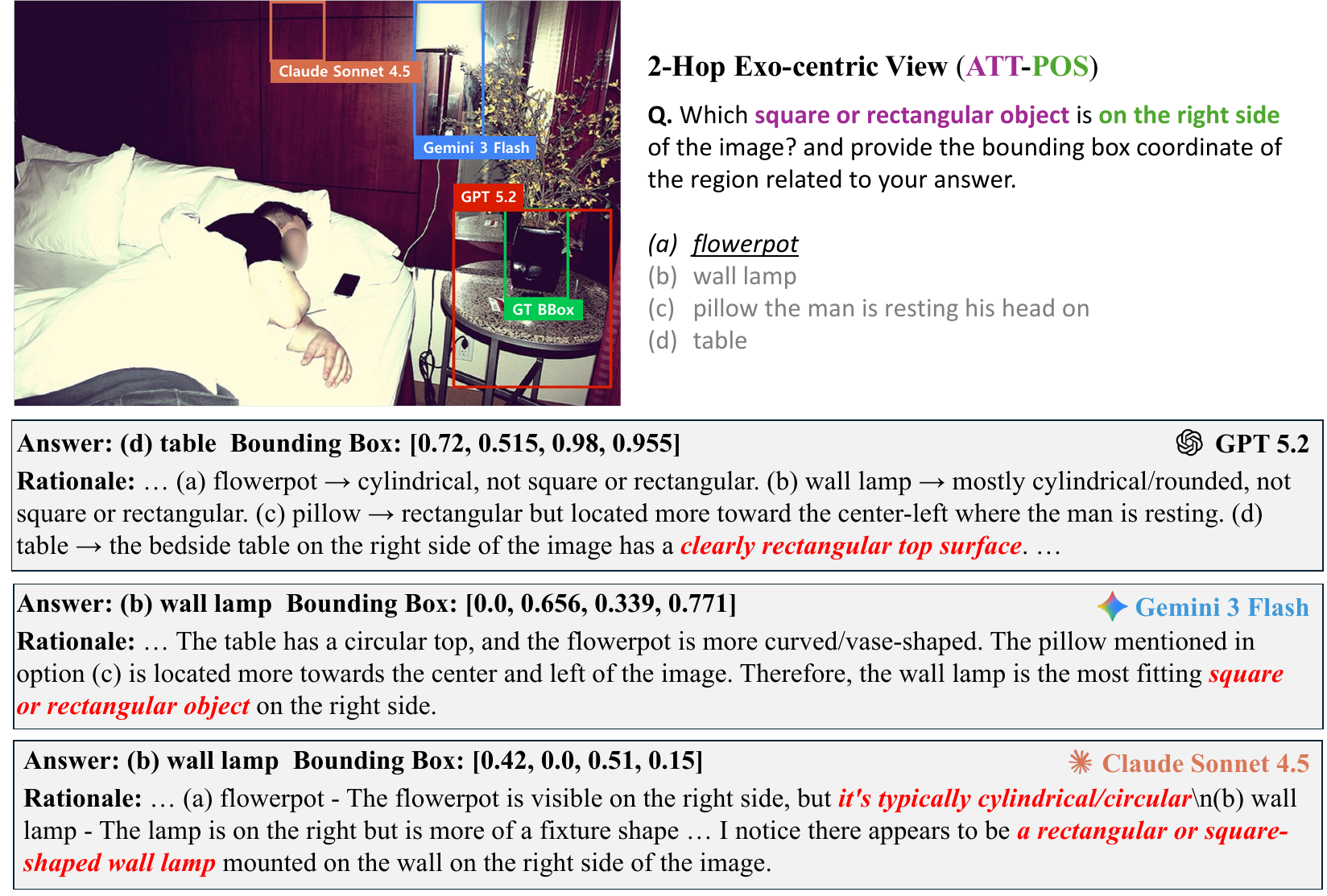}
    \caption{\textbf{Qualitative failure analysis on a 2-hop exo-centric question.} All three models fail to recognize the attribute condition.}
    \label{fig:qual_exo2hop3}
\end{figure}

\begin{figure}[t]
    \centering
        \includegraphics[width=\linewidth]{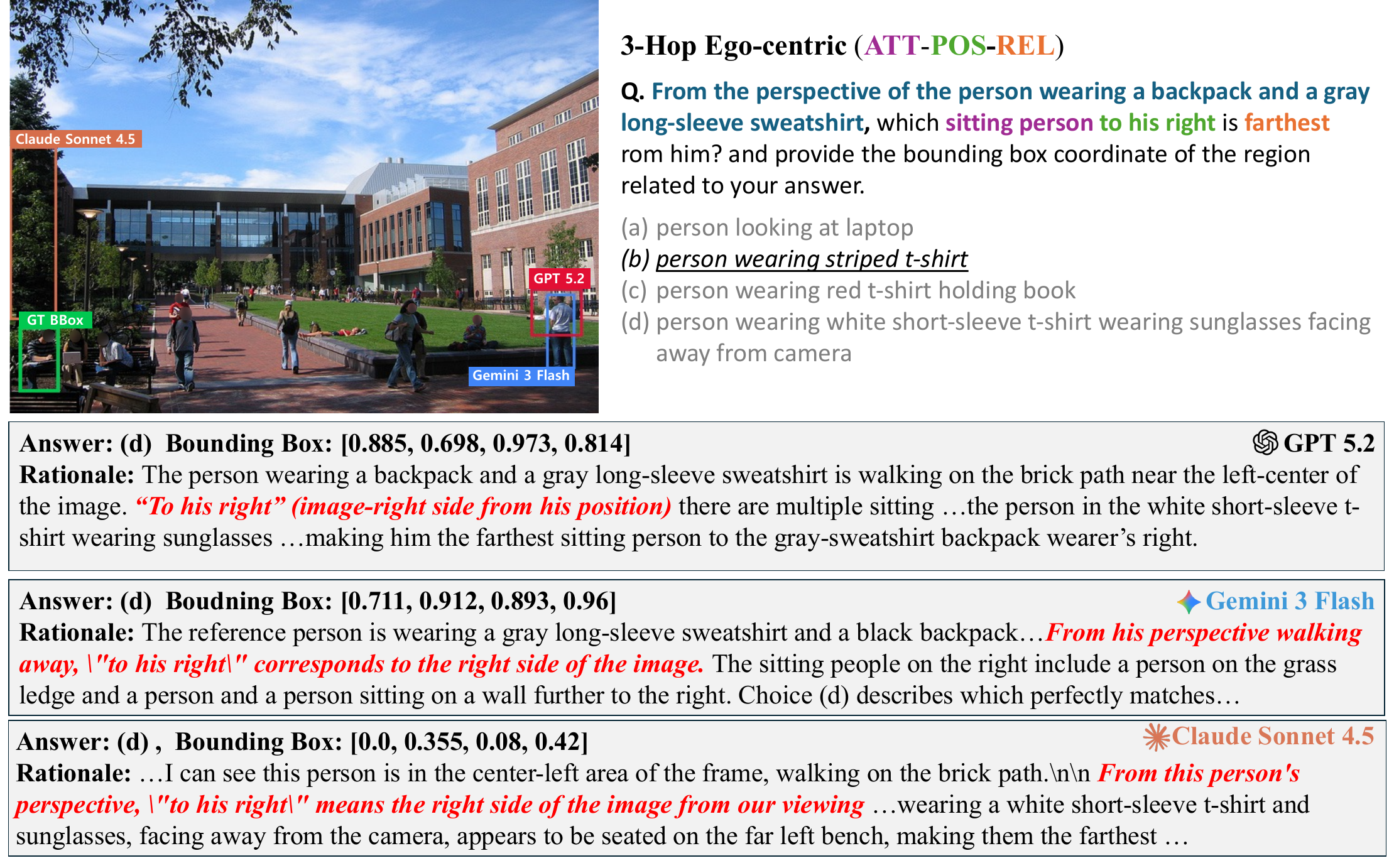}
    \caption{\textbf{Qualitative failure analysis on a 3-hop ego-centric question.} All three models fail to correctly identify the perspective.}
    \label{fig:qual_ego3hop1}
\end{figure}

\begin{figure}[t]
    \centering
        \includegraphics[width=\linewidth]{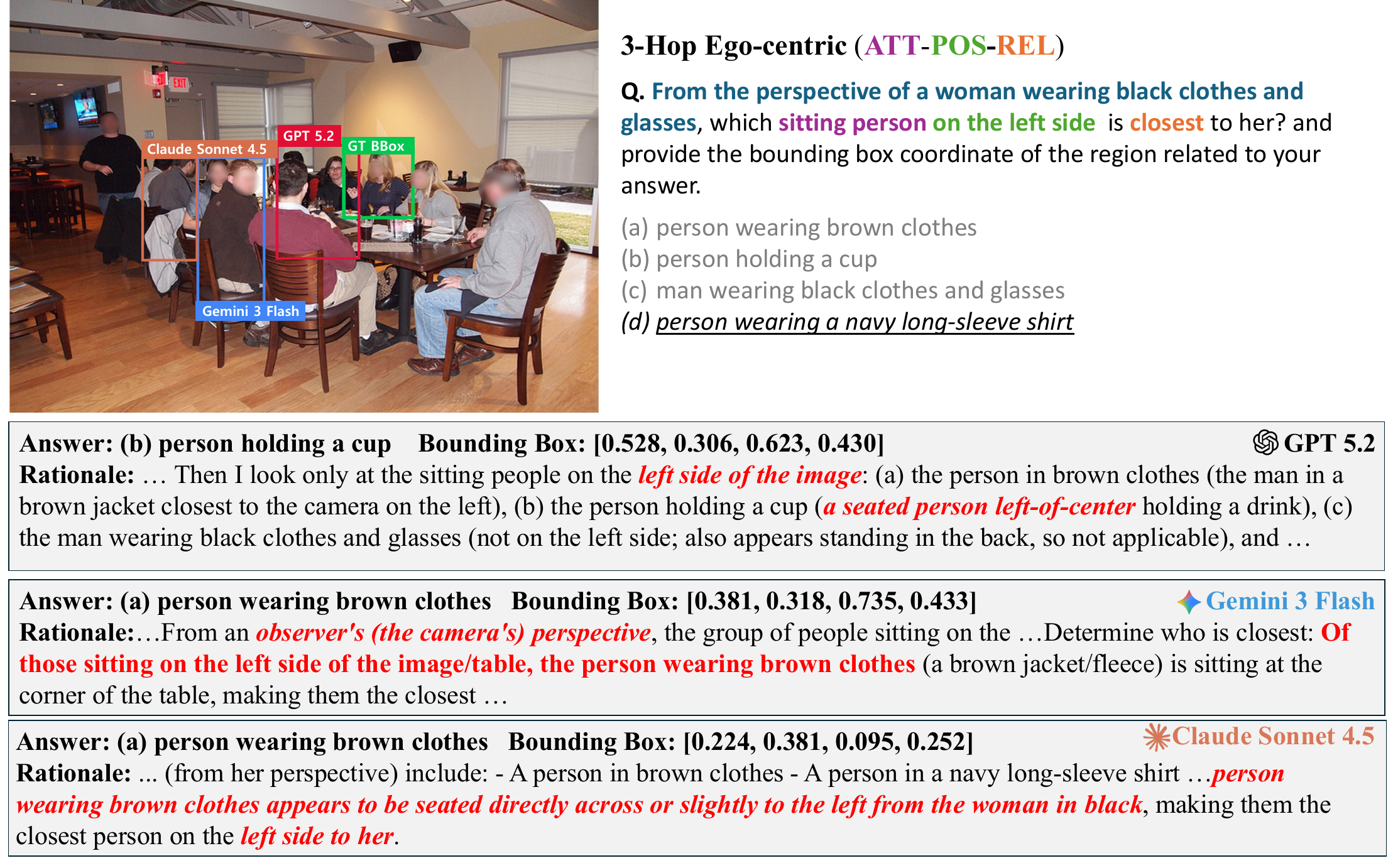}
    \caption{\textbf{Qualitative failure analysis on a 3-hop ego-centric question.} All three models fail to correctly identify the perspective.}
    \label{fig:qual_ego3hop2}
\end{figure}

\begin{figure}[t]
    \centering
        \includegraphics[width=\linewidth]{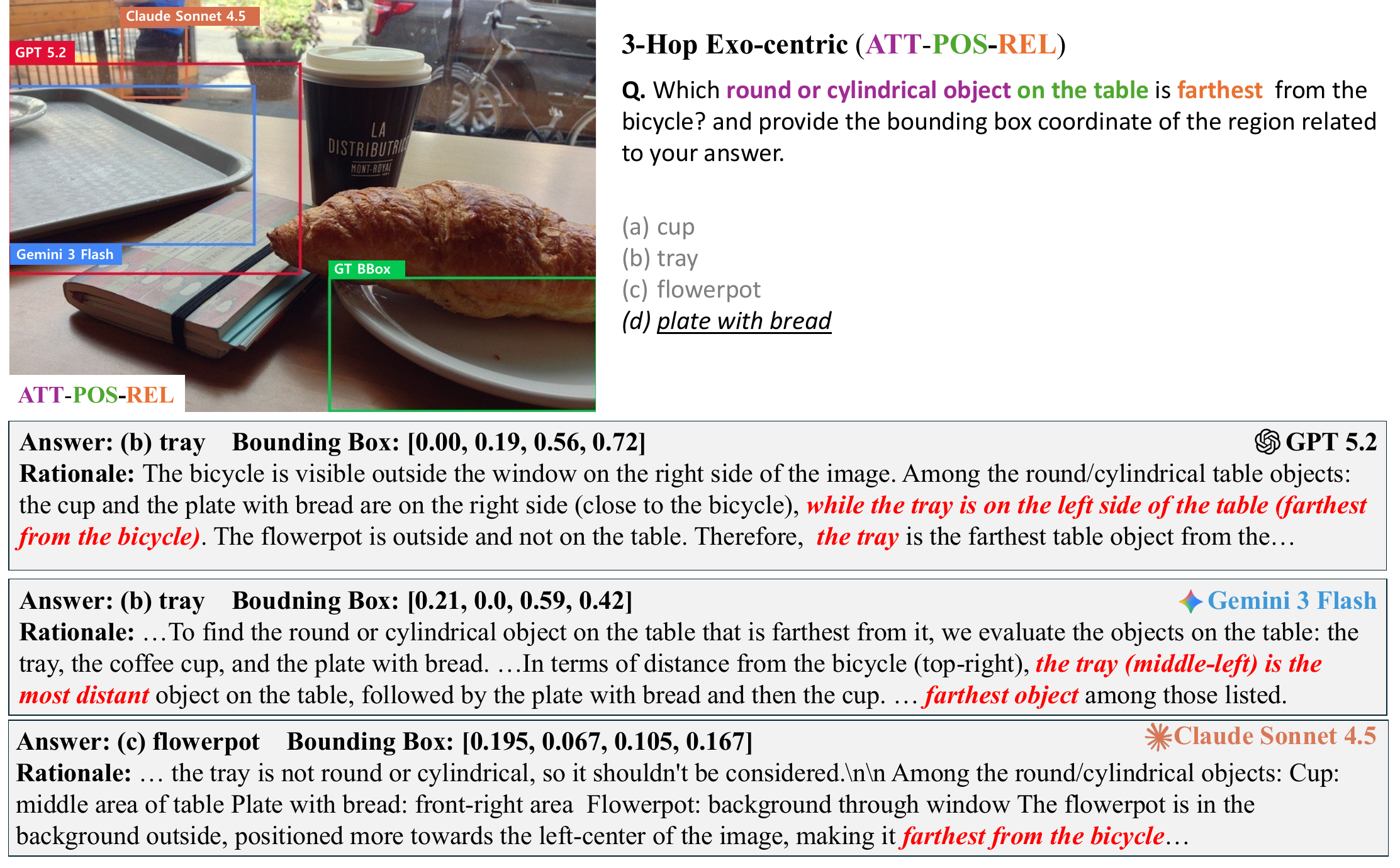}
\caption{\textbf{Qualitative failure analysis on a 3-hop exo-centric question.} GPT and Gemini fail on attribute condition recognition, whereas Claude fails on position condition recognition.}
    \label{fig:qual_exo3hop1}
\end{figure}

\begin{figure}[t]
    \centering
        \includegraphics[width=\linewidth]{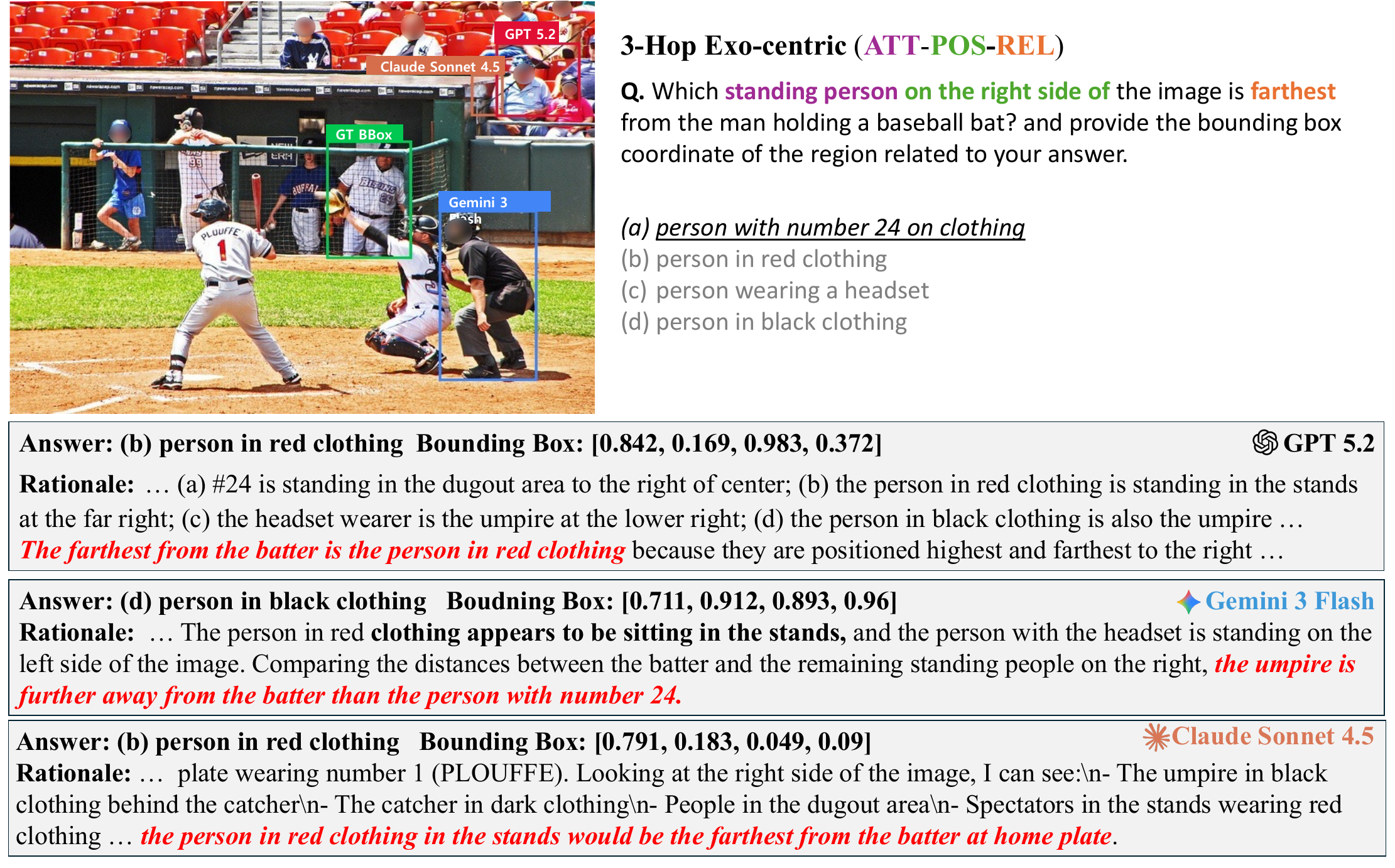}
\caption{\textbf{Qualitative failure analysis on a 3-hop exo-centric question.} GPT and Claude fail to recognize the attribute condition, whereas Gemini fails on relation recognition.}
    \label{fig:qual_exo3hop2}
\end{figure}

We provide additional qualitative results illustrating failure cases on 2-hop and 3-hop ego-centric and exo-centric questions. The results in~\cref{fig:qual_ego3hop1} and~\cref{fig:qual_ego3hop2} indicate that the models struggle even with basic perspective-taking, while~\cref{fig:qual_exo3hop1} and~\cref{fig:qual_exo3hop2} reveal that they often fail to retain intermediate conditions, such as attributes or positions, throughout the reasoning process. \cref{fig:qual_ego2hop1,fig:qual_ego2hop2,fig:qual_ego2hop3,fig:qual_exo2hop1,fig:qual_exo2hop2,fig:qual_exo2hop3} further present representative 2-hop failure cases covering all three pairwise compositions (\textsc{att+pos}, \textsc{att+rel}, and \textsc{pos+rel}). Compared with the 3-hop examples, the shorter reasoning chains are generally better preserved until the final prediction. However, the models still fail in different ways, even on the same sample. Overall, these examples suggest that the difficulty does not stem only from long reasoning chains. An error in a single intermediate cue can easily propagate and lead to an incorrect final answer. This again highlights the importance of evaluating compositional spatial reasoning beyond single-hop queries.

\clearpage

\vspace{-2cm}
\vspace{-2cm}

    
    
    
    
    
\vspace{-1cm}

    
    
    
    
    


\clearpage


\definecolor{ClaudeBG}{HTML}{FDF0E7}

\definecolor{ClaudeText}{HTML}{B46A3C}


\definecolor{GPTBG}{HTML}{EAF8EF}

\definecolor{GPTText}{HTML}{2E8B57}


\definecolor{GeminiBG}{HTML}{EAF2FF}

\definecolor{GeminiText}{HTML}{1F5FBF}

\definecolor{HeaderBG}{HTML}{D9EAF7}

\definecolor{BBoxBG}{HTML}{F8F8F8}


\newcolumntype{L}[1]{>{\RaggedRight\arraybackslash}p{#1}}

\newcolumntype{Y}{>{\RaggedRight\arraybackslash}X}


\newcommand{\modelClaude}{\cellcolor{ClaudeBG}\textcolor{ClaudeText}{\textbf{Claude Sonnet 4.5}}}

\newcommand{\modelGPT}{\cellcolor{GPTBG}\textcolor{GPTText}{\textbf{GPT-5.2}}}

\newcommand{\modelGemini}{\cellcolor{GeminiBG}\textcolor{GeminiText}{\textbf{Gemini3.0-Flash}}}


\newcommand{\tracefont}{\scriptsize}

\newcommand{\tracecaption}[2]{Full reasoning traces from #2 for the example shown in Fig.~\ref{#1}.}

\begin{table*}[t]
\centering
\fontsize{5.5pt}{6.2pt}\selectfont
\caption{\tracecaption{fig:failure_case}{GPT-5.2, Gemini3.0-Flash, and Claude Sonnet 4.5}}

\label{tab:sup_trace_rect}

\setlength{\tabcolsep}{3pt}

\renewcommand{\arraystretch}{0.95}

\begin{tabularx}{\textwidth}{L{0.14\textwidth} Y}

\toprule

\textbf{Model} & \textbf{Reasoning Trace} \\

\midrule

\modelGPT & \tracefont The sitting person is at the table with a gray laptop directly in front of them, and a black cellphone lying on the table closer to the camera. The framed picture is on the back wall above and behind the person, and the white door is also on the back/right wall. Among these square or rectangular objects, the one farthest from the sitting person is the wall-mounted framed picture, since it is clearly behind them and farther than the nearby door and table objects.\par\smallskip

Answer: (a) framed picture with a drawing of a person\par\smallskip

Bounding Box: \{\texttt{"bbox\_2d"}: [0.364, 0.035, 0.471, 0.190]\}\\

\midrule

\modelGemini & \tracefont Based on the visual information in the image, the man is sitting at a table and facing toward the camera, as indicated by his face being visible and the back of his laptop's lid facing the viewer.\par\smallskip

From the man's perspective, the objects mentioned in the choices are positioned as follows: (d) the gray laptop is directly in front of him on the table, likely only a foot or two away; (b) the black cellphone is also on the table, farther from him than the laptop and closer to the camera; (a) the framed picture with a drawing of a person is on the wall directly behind him; and (c) the white door is located on a far wall in the background of the scene, significantly farther away from him in 3D space than any of the items on the table.\par\smallskip

Even though the door is technically behind the plane of his body, it is the most distant object from him among the choices given. In many visual question-answering tasks, ``in front of them'' is used broadly to describe objects visible in the scene or from the camera's vantage point facing the person.\par\smallskip

Answer: (c) white door\par\smallskip

Bounding Box: \{\texttt{"bbox\_2d"}: [0.273, 0.784, 0.571, 0.929]\}\\
\midrule
\modelClaude & \tracefont I need to identify which rectangular or square object in front of the sitting person is farthest from them.\par\smallskip

Let me analyze the objects mentioned in the choices: (a) framed picture with a drawing of a person, visible on the shelf in the background; (b) black cellphone, on the table in front of the person; (c) white door, visible on the right side of the image; and (d) gray laptop, on the table directly in front of the person.\par\smallskip

From the person's perspective sitting at the table, the laptop is closest because it is right in front of them. The cellphone is on the table and relatively close. The framed picture is on the shelf above and behind the fireplace, while the white door is on the right wall.\par\smallskip

The framed picture appears to be the farthest object from the sitting person among the rectangular or square objects listed. It is located on a shelf in the background, above the fireplace.\par\smallskip

Answer: (a) framed picture with a drawing of a person\par\smallskip

Bounding Box: \{\texttt{"bbox\_2d"}: [0.3031, 0.0669, 0.4031, 0.1736]\} \\

\bottomrule

\end{tabularx}

\end{table*}

\end{document}